\documentclass[journal,twoside]{IEEEtran}

\IEEEoverridecommandlockouts
\usepackage{cite}
\usepackage{amsmath,amssymb,amsfonts}
\usepackage{graphicx}
\usepackage{textcomp}
\usepackage{algorithm,algorithmicx,algpseudocode}
\usepackage{booktabs}
\usepackage{mathtools}
\usepackage{subfig}
\usepackage{multicol}
\usepackage{arydshln}
\usepackage[usenames,dvipsnames,svgnames,table]{xcolor}
\definecolor{darkgreen}{rgb}{0,0.6,0}
\usepackage[bookmarks=true,colorlinks=true,pdfpagemode=UseNone,citecolor=darkgreen,linkcolor=black,urlcolor=BrickRed]{hyperref}
\usepackage{balance}
\usepackage{float}

\graphicspath{{plots/}}

\begin{document}

\title{A Replay-Constrained Simulation Framework for Personalization of Powered Knee--Ankle Prosthesis Controllers}

\author{Duong~Le,
        Ryan~Posh,
        Shihao~Cheng,
        Maani~Ghaffari,
        and~Robert~D.~Gregg,%
\thanks{This work was supported in part by the Vingroup Science and Technology Scholarship Program (D.~Le); the U.S. National Institutes of Health (NIH), National Institute of Child Health and Human Development (NICHD), under Award Number 1F32HD116414-01A1 (R.~Posh); and the NIH NICHD under Award Number R01HD094772 (S.~Cheng and R.~D.~Gregg). The content is solely the responsibility of the authors and does not necessarily represent the official views of the NIH.}%
\thanks{The authors are with the College of Engineering, University of Michigan, Ann Arbor, MI 48109, USA (e-mail: duongqle@umich.edu).}%
}

\markboth{Preprint}%
{Le \MakeLowercase{\textit{et al.}}: Replay-Constrained Simulation Framework for Personalization of Powered Knee-Ankle Prosthesis Controllers}

\maketitle

\begin{abstract}
Personalization of impedance controllers for powered prosthetic legs is critical to accommodating individual gait biomechanics but remains challenging. Existing methods rely on time-intensive human-in-the-loop exploration and/or constrain optimization to low-dimensional, single-joint parameter subspaces. Sim-to-real transfer has enabled high-dimensional locomotion control for legged robots, but in assistive device control the human partner remains un-modelable. We present a replay-constrained simulation framework: a MuJoCo-based simulator reproduces prosthetic knee-ankle dynamics while replaying recorded hip kinematics and feedback-based ground reaction forces from individual walking data, bypassing the need to model complex human neuromuscular control mechanisms. We demonstrate the framework with a deep reinforcement learning policy that personalizes phase-dependent stiffness, damping, and equilibrium angle at both joints simultaneously, maximizing a biomimicry-based reward computed solely from onboard prosthesis measurements. Experiments with three participants with transfemoral amputation during level-ground walking at 0.8~m/s demonstrate strong simulation-to-hardware predictive validity (Pearson $r=0.96$--$0.997$). The best-performing policy on hardware was consistently predicted within the top five simulation policies for all participants. The learned controllers improved overall biomimicry rewards by 42--59\% relative to the unpersonalized baseline. The framework supports scalable high-dimensional personalization of powered prosthetic legs and is amenable to extension to higher-dimensional controller parameterizations such as neural-network controllers.
\end{abstract}


\IEEEpeerreviewmaketitle

\section{Introduction}

Limb loss affects approximately 2.3 million individuals in the United States, projected to double to 5.6 million by 2060~\cite{rivera2024estimating}. While passive and microprocessor-controlled prostheses remain most commonly prescribed, they impose substantially increased metabolic cost and reduced self-selected walking speed relative to able-bodied individuals~\cite{russell2018influence}. Powered knee-ankle prostheses can address these limitations through improved gait biomechanics~\cite{best2023data,zhou2025comparing,Keller2026}, reduced metabolic demand~\cite{jayaraman2018impact, au2009powered}, and net-positive joint work~\cite{sup2009self, lawson2014robotic, elery2020design}, but realizing these benefits depends on control strategies personalized to user-specific anatomy, socket fit, alignment, and compensatory patterns~\cite{tucker2015control, gehlhar2023review, reznick2025clinical}. The inability to efficiently personalize control remains a significant barrier to clinical adoption.

Impedance control~\cite{hogan1985impedance} has emerged as the dominant framework for powered knee-ankle prosthesis control, leveraging spring-damper-like analogs to modulate joint torque as a function of angle and velocity~\cite{tucker2015control}. Empirical studies demonstrate that human ankle walking dynamics can be well-described by impedance models~\cite{hogan1985impedance, rouse2014estimation, lee2016summary}. Early finite-state machine (FSM) controllers used piecewise-constant impedance parameters for discrete states of gait~\cite{sup2009self, lawson2014robotic, simon2014configuring, eilenberg2010control}, enabling pioneering achievements including level walking, stair ascent, and running~\cite{lawson2014robotic, shultz2014running}. These FSM controllers, however, required extensive manual tuning ~\cite{simon2014configuring, best2023data} and exhibited discrete torque transitions that could compromise stability~\cite{thatte2019real, posh2023finite}. Variable impedance controllers parameterized by a continuous phase variable~\cite{anil2022control, best2023data, posh2024hybrid, best2025decoupling} can address the torque discontinuity by ensuring smooth continuous control, and data-driven parameterizations can further reduce the need for manual tuning. In particular, the data-driven Hybrid Kinematic-Impedance Controller (HKIC) parametrizes stance-phase impedance as third-order polynomial functions of gait phase~\cite{best2023data}, learned from able-bodied reference data for walking across speeds and inclines. Hardware validation with transfemoral amputee participants demonstrated that HKIC met or exceeded FSM performance in seven of eight biomechanical metrics~\cite{best2023data}, while enabling automatic task adaptation without manual tuning~\cite{best2023data,cheng2025ambilateral}. Despite these advantages, HKIC largely remains a population-level controller that does not account for user variance in biomechanics and movement strategies~\cite{reznick2025clinical}.

Recent approaches have attempted to personalize stance-phase impedance control but face fundamental limitations on hardware. Manual tuning achieves personalization within approximately 20 minutes~\cite{reznick2025clinical}, but exposes users to suboptimal controllers over 8--13 iterations and depends on subjective clinician observation. Human-in-the-loop reinforcement learning (HIL-RL) automates personalization from user feedback~\cite{wen2019online, li2021toward, gao2021reinforcement, wu2022reinforcement, alili2023novel}, including recent bilevel inverse-RL approaches for knee prostheses~\cite{liu2025addressing}, but remains constrained to discrete FSM parameterizations (12 parameters for knee-only prostheses) and still requires iterative on-hardware exploration. Classical optimization methods such as CMA-ES~\cite{zhang2017human} and Bayesian optimization~\cite{ding2018human} face the same hardware-iteration bottleneck and have been validated only on low-dimensional parameter spaces. Extending any of these to continuous phase-dependent multi-joint controllers such as HKIC~\cite{best2023data} or to neural-network controllers~\cite{kim2022seamless, nuesslein2024deep} substantially expands the parameter space and renders direct HIL training infeasible.

Hong \emph{et~al.}~\cite{hong2023towards, hong2025offline} addressed scalability by using principal component analysis (PCA) to reduce knee stiffness from 4 to 2 dimensions for surrogate-model-based RL, but this required three scope reductions: knee-only personalization, fixing damping and equilibrium angle at population values, and constraining solutions to the able-bodied PCA subspace, potentially excluding amputee-specific strategies. No existing method enables full-dimensional, multi-joint continuous impedance personalization that simultaneously optimizes stiffness, damping, and equilibrium angle at both knee and ankle without dimensionality reduction.

\begin{figure*}
  \centering
  \includegraphics[width=\textwidth]{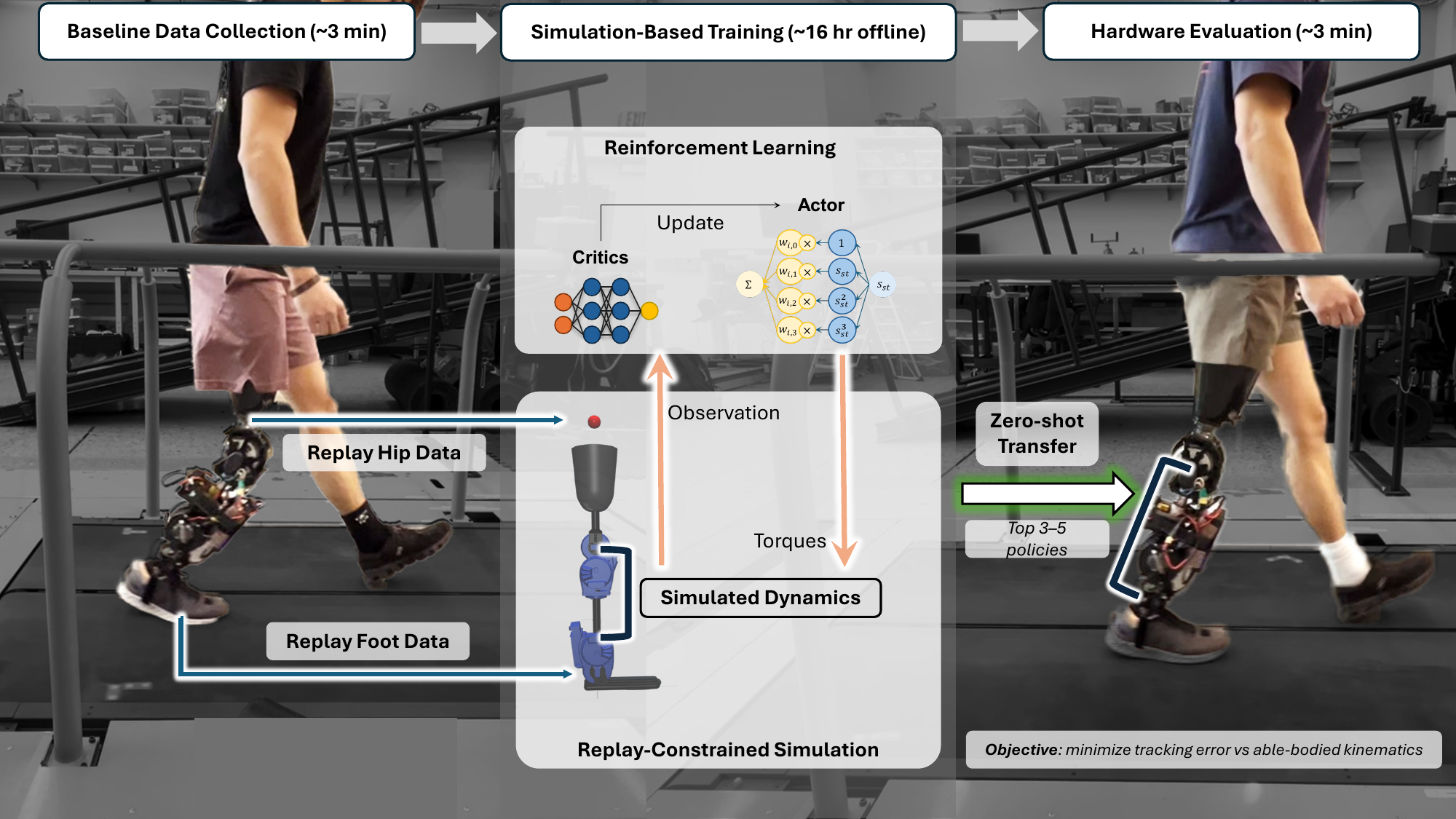}
  \caption{\textbf{Replay-constrained personalization framework for stance-phase impedance control.}
  \textbf{(Left)} Baseline data collection: three minutes of treadmill walking at 0.8~m/s yields hip kinematics, ground reaction forces, and foot orientation data from onboard prosthesis sensors.
  \textbf{(Center)} Simulation-based training: the replay-constrained simulator replays boundary conditions (hip kinematics, feedback-augmented ground reaction forces) while simulating knee-ankle dynamics under impedance control; TD3~\cite{fujimoto2018addressing} optimizes phase-dependent impedance coefficients over approximately 16--29 hours of offline computation (equivalent to approximately 15--26 hours of on-prosthesis walking avoided).
  \textbf{(Right)} Hardware evaluation: top 3--5 simulation-ranked policies are deployed zero-shot for prosthetist assessment (approximately 3 minutes total).
  Objective: minimize joint angle tracking error relative to able-bodied kinematics.}
  \label{fig:overview}
\end{figure*}

While these simulation-based RL challenges remain for many assistive devices, the broader robotics community has achieved zero-shot sim-to-real transfer for high-dimensional locomotion control~\cite{rudin2022learning, radosavovic2024humanoid}. These successes rely on a critical assumption: the robot's dynamics are fully characterized by the physics simulator, and no unmodeled adaptation occurs in the loop. In assistive device applications, however, control requires coordinating two intelligent agents, the device and the user, whose coupled dynamics cannot be captured by simulating the device alone. The user actively adapts their movement strategy in response to device behavior; for example, persons with transfemoral amputation modulate hip extension moments, trunk flexion, and step length in response to prosthetic knee alignment changes~\cite{koehler2016biomechanical}. Consequently, policies optimized against a fixed or predicted human model may fail when the real user adapts differently than expected.

Despite this fundamental challenge, recent work has applied musculoskeletal simulation to train end-to-end assistive device controllers that transfer zero-shot to hardware~\cite{luo2024experiment, park2026learning}. These approaches, however, target population-level controller design rather than personalizing an existing controller to an individual user's biomechanics. Data-driven personalization methods can optimize user-specific assistance but still require extended on-hardware walking~\cite{slade2022personalizing}. Simulation-based personalization has been explored for powered knee prostheses through bilevel inverse RL~\cite{liu2022irl} and collaborative multi-agent RL that explicitly models human adaptation~\cite{wu2022collaborative}, but both remain restricted to knee-only FSM parameterizations and lack hardware validation with amputee participants. Wang \emph{et~al.}~\cite{wang2025hitl} further demonstrated the difficulty of predicting individual human responses, finding discrepancies between simulated and real participants in the adaptation strategies that emerged during walking with the device. These limitations motivate simulation strategies that encode measured individual behavior rather than attempting to predict it.

Therefore, we present a \emph{replay-constrained simulation framework} for high-dimensional stance-phase impedance personalization of powered knee-ankle prostheses, which we demonstrate by training a reinforcement learning (RL) policy that personalizes HKIC, originally learned from able-bodied reference data~\cite{best2023data}, to each amputee user during level-ground walking (Fig.~\ref{fig:overview}). The framework makes this possible without requiring off-board sensors, exposing users to suboptimal controllers, or demanding users to complete extended hardware walking sessions, ensuring that this high-dimensional personalization remains deployable. Direct hardware-based RL is infeasible at this scale: optimizing the full continuous phase-dependent impedance trajectory would require thousands of controller evaluations and prolonged hardware walking sessions, exposing users to poorly-performing controllers during the search. Instead, we sidestep the need for explicit human modeling by constructing a MuJoCo-based simulator that replays user-specific hip kinematics and proportional-feedback-augmented ground reaction forces from a baseline HKIC walking trial, while simulating knee-ankle dynamics under learned impedance control. Paired with Twin Delayed Deep Deterministic Policy Gradient (TD3)~\cite{fujimoto2018addressing}, this framework simultaneously personalizes continuous phase-dependent stiffness, damping, and equilibrium angle at both knee and ankle joints without dimensionality reduction or surrogate modeling. TD3 is chosen as a representative gradient-based RL method; such methods extend naturally to higher-dimensional controller parameterizations such as neural-network controllers~\cite{kim2022seamless, nuesslein2024deep}, where gradient-free alternatives such as evolution strategies do not scale. The framework itself is optimizer-agnostic and could equally be paired with classical methods on the structured polynomial parameterization studied here.

In this work, experiments with three transfemoral amputee participants validate that this framework (i) achieves strong sim-to-real predictive validity without requiring neuromuscular models, (ii) enables continuous phase-dependent personalization of multi-joint impedance without dimensionality reduction, and (iii) establishes a safe clinical workflow (Fig.~\ref{fig:overview}) that eliminates user exposure to suboptimal controllers and extended hardware walking time. The hardware-best policy was predicted by the RL simulation environment within the top five simulations out of over 5,000 simulation-ranked candidates for all participants, and learned controllers substantially improved biomechanical performance relative to the HKIC baseline. Importantly, simulation-trained controllers enabled stable hardware fine-tuning, whereas naïve human-in-the-loop RL initialized from the unpersonalized HKIC baseline resulted in performance degradation for two of three participants, demonstrating the practical necessity of simulation pretraining for safe, high-dimensional impedance personalization. Together, these results establish replay-constrained simulation as a practical pathway toward scalable, high-dimensional assistive device personalization that minimizes user burden while maintaining clinical safety and viability.

\section{Hardware and Baseline Control}
\label{sec:hardware-baseline}

This work builds upon established hardware and control architectures to isolate the contributions of the replay-constrained simulation framework for impedance personalization. We employ the powered knee--ankle prosthesis platform developed by Elery \emph{et~al.}~\cite{elery2020design} and the HKIC developed by Best \emph{et~al.}~\cite{best2023data} in order to implement our personalization framework. This section briefly describes the experimental hardware (Section~\ref{sec:hardware}) and the baseline HKIC controller (Section~\ref{sec:hkic-baseline}) that provides both the comparison baseline and the initialization point for personalization.

\subsection{Experimental Hardware and Sensing}
\label{sec:hardware}

We conducted all experiments using the powered knee--ankle prosthesis with high-torque, low-impedance actuators developed in~\cite{elery2020design}. The device employs brushless DC motors with low-reduction planetary transmissions at each joint. Optical encoders mounted on motor shafts measure joint positions, a six-axis load cell distal to the ankle measures ground reaction forces and moments, an inertial measurement unit on the thigh provides orientation and angular velocity estimates for real-time phase estimation, and a foot-mounted IMU measures foot orientation for contact type determination during simulation (Section~\ref{sec:replay-architecture}). Complete hardware specifications are detailed in~\cite{elery2020design, cheng2025ambilateral}.

\subsection{Hybrid Kinematic-Impedance Control}
\label{sec:hkic-baseline}

This personalization framework builds upon the HKIC developed in~\cite{best2023data}, which provides continuous phase-dependent impedance control derived from able-bodied biomechanical data. In this work, HKIC serves as both the baseline controller for performance comparison and the initialization point for RL. The controller operates in a hybrid manner: the stance phase employs impedance control to regulate compliant ground interaction, while the swing phase uses position control for kinematic tracking. This subsection summarizes the stance-phase impedance parameterization that our RL framework personalizes.

\subsubsection{Stance-Phase Impedance Control}
\label{sec:hkic-stance}

During stance, each joint commands torques according to a continuous, phase-varying impedance law:
\begin{equation}
\begin{aligned}
\tau_j(t) &= K_j(s_{\mathrm{st}})\,\big(\theta_j(t)-\theta^{\mathrm{eq}}_j(s_{\mathrm{st}})\big) \\
          &\quad + B_j(s_{\mathrm{st}})\,\dot{\theta}_j(t), \qquad j\in\{\text{knee},\text{ankle}\},
\end{aligned}
\label{eq:hkic_stance}
\end{equation}
where $s_{\mathrm{st}}(t) \in [0,1]$ is the stance-phase variable, initialized at heel strike and advancing monotonically until toe-off during steady walking. The stiffness $K_j(s_{\mathrm{st}})$, damping $B_j(s_{\mathrm{st}})$, and equilibrium angle $\theta^{\mathrm{eq}}_j(s_{\mathrm{st}})$ are continuous functions of stance phase derived from able-bodied walking data. All commanded torques are rate-limited and saturated to hardware-safe bounds before actuation.

\subsubsection{Phase-Dependent Impedance Parameters}
\label{sec:hkic-parameters}

HKIC represents each impedance parameter as a third-order polynomial in stance phase:
\begin{equation}
p_i(s_{\mathrm{st}}) = \sum_{k=0}^{3} w_{i,k}\, s_{\mathrm{st}}^k, \quad i \in \{1,\ldots,6\},
\label{eq:hkic_polynomial}
\end{equation}
where $p_i \in \{K_{\text{knee}}, B_{\text{knee}}, \theta^{\mathrm{eq}}_{\text{knee}}, K_{\text{ankle}}, B_{\text{ankle}}, \theta^{\mathrm{eq}}_{\text{ankle}}\}$ and $w_{i,k}$ are polynomial coefficients optimized from able-bodied data to match population-average kinetic trends across level-ground walking. The third-order polynomial basis provides sufficient flexibility to capture physiological impedance variation throughout stance while maintaining smooth, continuous trajectories.

\subsubsection{Phase Estimation and Gait Event Detection}
\label{sec:phase-estimation}

The stance-phase variable $s_{\mathrm{st}}(t)$ used in~\eqref{eq:hkic_stance} and~\eqref{eq:hkic_polynomial} is derived from a gait-cycle phase variable $s(t)$, computed using a piecewise monotonic algorithm applied to thigh orientation as measured by the thigh-mounted IMU~\cite{best2023data}. This algorithm includes stride-by-stride adaptation of thigh trajectory features and phase linearization parameters, enabling accommodation of individual gait patterns within approximately 5--20 strides without manual tuning. Heel strike and toe-off are detected via vertical ground reaction force thresholds from the distal load cell. 
The stance-phase variable is then obtained as $s_{\mathrm{st}}(t) = s(t)/\bar{s}_{\mathrm{TO}}$, where $\bar{s}_{\mathrm{TO}}$ is the average value of $s(t)$ at toe-off.

\section{Replay-Constrained Simulation Framework}
\label{sec:digital-twin}

This section presents a replay-constrained simulator that enables policy optimization without requiring extensive on-hardware exploration or exposing users to poor-performing controllers. The approach partitions the human-prosthesis system into measured boundary conditions (hip kinematics and proportional-feedback-augmented ground reaction forces from baseline walking) and simulated prosthesis dynamics (knee-ankle impedance control), avoiding the need for accurate neuromuscular models of residual limb dynamics. The key assumption is that hip kinematics and foot orientation remain approximately stable, varying substantially less than the directly controlled knee and ankle joints, when stance-phase knee-ankle impedance is perturbed within physiologically reasonable bounds. This assumption is empirically validated for the walking condition studied here in Section~\ref{sec:disc-sim-to-real}, with per-subject numerical values reported in Table~\ref{tab:supp-replay-validation} and Fig.~\ref{fig:supp-replay-validation}.

\subsection{Data Collection for Simulator Construction}
\label{sec:data-collection}

Baseline walking data for constructing the replay-constrained simulator were collected as described in Section~\ref{sec:experimental-protocol}. Each participant walked at 0.8~m/s for three minutes under HKIC baseline control, yielding approximately 100 strides after processing. Recorded sensor data included joint encoder measurements (knee, ankle angles and velocities), IMU measurements (thigh and foot orientation and angular velocity), load cell measurements (ground reaction forces and moments), and gait phase estimates. Strides were normalized to 150 samples (90 stance, 60 swing) via phase-based resampling for simulator replay. Critically, all recorded sensor data were obtained from onboard prosthesis instrumentation, eliminating the need for external motion capture infrastructure. This design choice enables clinical reproducibility: the personalization framework can be deployed in standard prosthetics clinics without specialized biomechanics laboratories. Per-subject collection is essential to encoding individual differences in residual-limb dynamics, socket fit, and gait patterns, seeding the personalization workflow.

\subsection{Replay-Simulation Architecture and Implementation}
\label{sec:replay-architecture}

The simulator partitions the human-prosthesis system into replayed boundary conditions and simulated impedance dynamics. Hip motion (horizontal translation, vertical translation, and sagittal rotation) is commanded kinematically from baseline walking data, ground reaction forces are applied with proportional-feedback augmentation, and knee and ankle joints evolve dynamically under impedance control with torques generated by the learned policy.

\subsubsection{Hip Kinematics Calculation from Contact Constraints}
Hip translational trajectories are derived from baseline joint angle data through forward kinematics with contact constraints, avoiding direct reliance on hip IMU measurements. For each timestep, the hip position is calculated as:
\begin{equation}
\mathbf{p}_{\mathrm{hip}} = \mathbf{p}_{\mathrm{contact}} - \mathbf{p}_{\mathrm{contact}}^{\mathrm{hip}},
\label{eq:hip-ik}
\end{equation}
where $\mathbf{p}_{\mathrm{contact}}$ is the world-frame position of the active contact point (heel or toe) and $\mathbf{p}_{\mathrm{contact}}^{\mathrm{hip}}$ is its position relative to the hip obtained through forward kinematics of the prosthesis kinematic chain. Contact type is determined from foot pitch sign measured by the foot-mounted IMU, with a binary heel/toe switch that introduces negligible discontinuity in the reconstructed hip trajectory. Per-phase contact-constraint definitions, the analogous hip-velocity computation, and the biomechanical rationale for the binary classification are detailed in the supplementary material (Section~\ref{sec:supp-hip-ik}).

\subsubsection{Ground Reaction Force Replay}
Ground reaction wrench components $w_i \in \{F_x, F_z, M_y\}$ from baseline data $w_{i,\mathrm{data}}$ are applied to a simulated load cell body located below the ankle joint~\cite{elery2020design}, matching the physical sensor location rather than the foot-ground contact point. These components are augmented with proportional feedback to maintain foot orientation:
\begin{equation}
w_i = \alpha_i w_{i,\mathrm{data}} + k_{p,i}(\phi_{\mathrm{ref}} - \phi_{\mathrm{sim}}),
\label{eq:grf-feedback}
\end{equation}
where $\alpha_i$ are mixing gains, $\phi$ denotes foot orientation, and $k_{p,i}$ are proportional feedback gains, provided in the supplementary material (Table~\ref{tab:supp-grf-gains}). Per-subject values are tuned to hold the simulated foot orientation to its reference while keeping the applied ground reaction forces close to the recorded loadcell data. These gains are then fixed during optimization.

\subsubsection{MuJoCo Prosthesis Model}
The user-prosthesis system is modeled in MuJoCo~\cite{todorov2012mujoco} as a five-degree-of-freedom kinematic chain: slider joints for hip translation, a revolute joint for hip rotation, and revolute joints for knee and ankle flexion. Link masses, inertias, and geometric parameters match the physical device~\cite{elery2020design, keller2023gait}, with segment lengths adjusted per participant (Table~\ref{tab:supp-participants}). Visualization meshes were adapted from the Open-Source Leg MuJoCo model~\cite{azocar2020design, tan2025myoassist}. During simulation, hip translational and rotational trajectories calculated via Eq.~\ref{eq:hip-ik} are commanded kinematically, while knee and ankle joints evolve dynamically under impedance control (Eq.~\ref{eq:hkic_stance}) with torques generated by the learned policy. Dynamics are integrated using MuJoCo's implicit Euler method with 1~ms timesteps.

\section{Controller Tuning Framework}
\label{sec:rl-framework}

This section describes how the replay-constrained simulator of Section~\ref{sec:digital-twin} is used to personalize HKIC. A composite per-stride reward, evaluated on simulated walking strides, drives the optimization of HKIC's polynomial impedance coefficients (Section~\ref{sec:hkic-parameters}), decoupling controller tuning from on-hardware iteration. Section~\ref{sec:action-policy} describes the action space and polynomial parameterization, Section~\ref{sec:reward} presents the composite reward design, and Section~\ref{sec:td3} describes the TD3 demonstration.

\subsection{Action Space and Polynomial Parameterization}
\label{sec:action-policy}

The action space $\mathcal{A} \subset \mathbb{R}^{6}$ comprises six instantaneous impedance values at each timestep: stiffness $K_j(s_{\mathrm{st}})$, damping $B_j(s_{\mathrm{st}})$, and equilibrium angle $\theta^{\mathrm{eq}}_j(s_{\mathrm{st}})$ for knee and ankle joints ($j \in \{\mathrm{knee}, \mathrm{ankle}\}$), evaluated at the current normalized stance phase $s_{\mathrm{st}} \in [0,1]$. The deterministic actor $\mu_\psi: [0,1] \to \mathbb{R}^{6}$ preserves the HKIC baseline's polynomial structure by parameterizing the action vector through the polynomial coefficient matrix $\mathbf{W}$:
\begin{equation}
\mathbf{a}(s_{\mathrm{st}}) = \mathbf{W}\boldsymbol{\phi}(s_{\mathrm{st}}),
\label{eq:polynomial_action}
\end{equation}
where $\boldsymbol{\phi}(s_{\mathrm{st}})$ is the polynomial basis from Eq.~\eqref{eq:hkic_polynomial}; the actor's trainable parameters $\psi$ correspond directly to $\mathbf{W}$, so the forward pass evaluates $\mu_\psi(s_{\mathrm{st}}) = \mathbf{W}\boldsymbol{\phi}(s_{\mathrm{st}})$. At each stance timestep, these impedance values are substituted into the HKIC stance impedance law~\eqref{eq:hkic_stance} to compute commanded joint torques. The actor receives only $s_{\mathrm{st}}$ as input, preserving HKIC's phase-dependent structure while enabling end-to-end gradient-based optimization. The actor is initialized with polynomial coefficients extracted from the HKIC baseline~\cite{best2023data}, biasing exploration toward physiologically plausible trajectories. All learning occurs in normalized action space; normalization bounds are in the supplementary material (Section~\ref{sec:supp-action-norm}, Table~\ref{tab:supp-limits}).

\subsection{Reward Function}
\label{sec:reward}

The reward function balances four complementary objectives to produce stance-phase impedance parameters that are biomimetic, physically realizable, and kinetically smooth; swing-phase timesteps are excluded since only stance impedance is optimized. The composite per-timestep reward is
\begin{equation}
r_t = \lambda_{\mathrm{base}} \sum_{j \in \{\mathrm{knee}, \mathrm{ankle}\}} \left( r^{\theta}_{j,t} + r^{\tau}_{j,t} + r^{\mathrm{sm}}_{j,t} + r^{\mathrm{damp}}_{j,t} \right),
\label{eq:reward_composite}
\end{equation}
where $\lambda_{\mathrm{base}} = 20.0$ is a multiplicative scaling chosen empirically for return-magnitude readability and learning-rate calibration; because it scales all four terms equally, it does not change the relative ranking of candidate controllers. $r^{\theta}_{j,t}$ represents the normative angle tracking reward, $r^{\tau}_{j,t}$ is the torque agreement reward, $r^{\mathrm{sm}}_{j,t}$ is the smoothness reward, and $r^{\mathrm{damp}}_{j,t}$ is the damping floor penalty, each described below. All error terms are normalized by the peak-to-peak range of able-bodied stance-phase data to ensure scale-invariant optimization across joints. Joint-specific weights are provided in the supplementary material (Table~\ref{tab:supp-reward-weights}) and detailed formulations are provided in the supplementary material (Section~\ref{sec:supp-reward}).

In this work, able-bodied gait serves as a normative benchmark rather than a prescriptive target for optimal gait in prosthesis users; the rationale for this choice and prior empirical support are detailed in the supplementary material (Section~\ref{sec:supp-reward}).

\subsection{Implementation: TD3}
\label{sec:td3}

Stance-phase transitions are treated as a finite-horizon MDP: at each timestep, the environment provides a 16-dimensional observation $o_t \in \mathcal{O} \subset \mathbb{R}^{16}$ comprising stance-phase progress, joint kinematics and commanded torques, IMU measurements, and ground reaction forces (Table~\ref{tab:supp-observation-space}); the actor outputs the six phase-dependent impedance values $a_t \in \mathcal{A} \subset \mathbb{R}^{6}$ from Eq.~\eqref{eq:polynomial_action}; and TD3~\cite{fujimoto2018addressing} maximizes the expected cumulative discounted return $\mathbb{E}[\sum_t \gamma^t r_t]$ over the composite reward $r_t$ from Eq.~\eqref{eq:reward_composite}. TD3 is selected as a representative gradient-based method whose neural-network function approximators scale to higher-dimensional policy classes such as neural-network controllers~\cite{kim2022seamless, nuesslein2024deep}, where gradient-free alternatives such as evolution strategies do not. Twin critic networks are implemented as multilayer perceptrons over the full observation paired with normalized actions, providing state-action value estimates for the TD3 update.

TD3 stabilizes continuous control through clipped double Q-learning with twin critics to mitigate overestimation bias, delayed policy updates to reduce gradient variance, and target-policy smoothing to improve robustness. Training is organized into episodes spanning multiple consecutive walking strides. During simulation training, zero-mean Gaussian noise is added to actor outputs to encourage exploration; during all hardware sessions, the actor outputs deterministic actions without exploration noise to ensure safe prosthesis behavior. Network architecture, training procedures, and hyperparameters are provided in the supplementary material (Section~\ref{sec:supp-td3}, Table~\ref{tab:supp-hyperparams}).

\section{Experimental Protocol}
\label{sec:experimental-protocol}

The experimental protocol comprised two supervised treadmill sessions per participant, conducted on separate days, to evaluate simulation-guided personalization of stance-phase impedance parameters for a powered knee-ankle prosthesis during level-ground walking at 0.8~m/s. Session~1 collected baseline data using the HKIC controller and conducted naïve HIL training to establish a comparison baseline for conventional RL.
Following Session~1, signal processing and stride partitioning prepared baseline data for training (Section~\ref{sec:data-prep}), after which simulation-based policy optimization proceeded for 5,000--9,000 episodes requiring approximately 16--29~hours without participant involvement.
Session~2 evaluated 25 simulation-selected candidate controllers on hardware through sequential deployment (one controller per episode), followed immediately by 25 episodes of HIL fine-tuning, yielding 50 on-hardware episodes total. Prior to data collection in each session, participants walked briefly to allow phase estimation parameters to stabilize (Section~\ref{sec:phase-estimation}). 
All procedures were approved by the University of Michigan Institutional Review Board (HUM00230065).

\subsection{Participants}
\label{sec:participant}

Three individuals with unilateral transfemoral amputation participated in this study (Fig.~\ref{fig:experimental_setup}; see supplementary material, Table~\ref{tab:supp-participants} for demographics). Participants were recruited based on Medicare Functional Classification Level K3 or K4, indicating community or high-activity ambulation capability. All participants had prior experience with microprocessor-controlled knee prostheses and were familiar with the powered prosthesis used in this study from previous research protocols.

\subsection{Session 1: Baseline and Naïve Training}
\label{sec:session1}

\subsubsection{Baseline Data Collection}
\label{sec:baseline-data}

Each participant walked continuously for three minutes at the fixed treadmill speed of 0.8~m/s using the HKIC baseline controller. All sensor streams (Section~\ref{sec:hardware}) were recorded at 6~ms timesteps for subsequent use in the replay-constrained simulation framework.

\subsubsection{Naïve Human-in-the-Loop Training}
\label{sec:naive-training}

After a short rest following baseline data collection, a 50-episode naïve training session was conducted to establish a comparison baseline for conventional HIL RL. Each participant walked at 0.8~m/s with seven strides per episode while TD3 optimized impedance parameters using the RL framework (Section~\ref{sec:rl-framework}) with hyperparameters specified in Table~\ref{tab:supp-hyperparams}. The gradient update rate was set to $u=10$ steps per episode (versus $u=1$ in simulation) to enable observable policy changes within the limited 50-episode session. Seated rest periods of self-selected duration were permitted after episodes 17 and 34 to minimize fatigue, during which data collection was paused; TF01 completed all 50 episodes continuously, while TF02 and TF03 utilized the scheduled rest periods.

\subsection{Simulation-Based Training}
\label{sec:simulation-training}

\subsubsection{Data Preparation}
\label{sec:data-prep}

Following Session~1, signal processing and stride normalization (Section~\ref{sec:data-collection}) were applied to each participant's three-minute baseline dataset, yielding approximately 100 usable strides per participant. Strides were shuffled using a fixed random seed and partitioned for simulation training, with 60 strides allocated to training environments and 40 strides to validation environments.

\begin{figure}[t]
\centering
\includegraphics[width=\columnwidth]{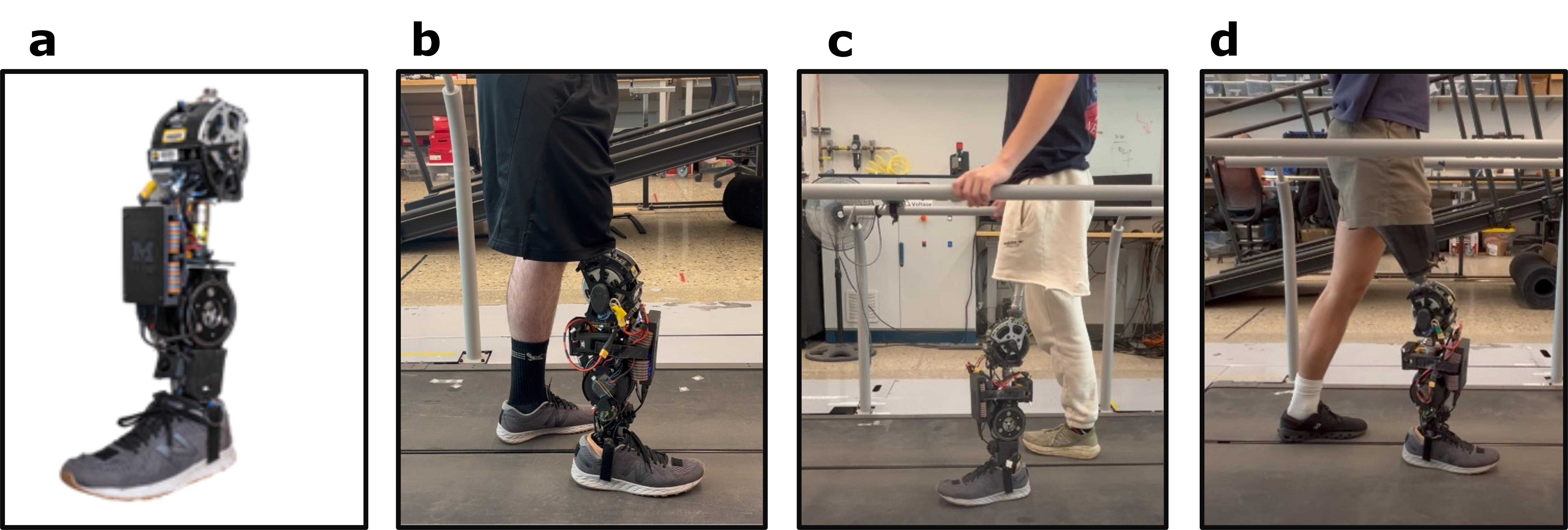}
\caption{Experimental hardware and participants. 
\textbf{(a)} Powered knee-ankle prosthesis. 
\textbf{(b--d)} Three participants with transfemoral amputation (TF01, TF02, TF03).}
\label{fig:experimental_setup}
\end{figure}

\subsubsection{Training Procedure}
\label{sec:training-procedure}

Policy optimization proceeded in the replay-constrained simulator without participant involvement, using the partitioned baseline dataset (Section~\ref{sec:data-prep}) and the RL framework detailed in Section~\ref{sec:rl-framework}. Training proceeded for 5,000--9,000 episodes (16--29~hours offline computation, equivalent to 15--26~hours of hardware trials avoided) on a workstation with NVIDIA GPU acceleration, employing vectorized parallel simulation with ten concurrent environment instances per episode (see supplementary material, Section~\ref{sec:supp-td3}). Validation performance was monitored throughout training.

\subsubsection{Candidate Controller Selection}
\label{sec:candidate-selection}

During training, models achieving new best validation performance were automatically saved, creating a sequence of checkpoints spanning the training trajectory. Twenty-five candidate controllers were selected per participant, with models drawn from both high-validation plateau regions and earlier training phases to probe simulation fidelity across varying policy quality levels (supplementary material, Section~\ref{sec:supp-protocol}).

\subsection{Session 2: Hardware Evaluation}
\label{sec:session2}

On a separate day, the participant returned for Session~2 to evaluate the simulation-selected candidate controllers on hardware through a continuous 50-episode session at 0.8~m/s. Episodes 1--25 deployed the 25 candidate controllers sequentially with frozen actor parameters and per-episode critic updates (critic warm-up), enabling zero-shot policy assessment while accumulating 25 episodes of critic experience to match the na\"ive baseline (Section~\ref{sec:naive-training}) at episode~26. Episodes 26--50 then conducted HIL fine-tuning with full TD3 updates ($u=10$, policy delay $d=2$), with the replay buffer containing all stance-phase transitions accumulated from episodes 1--50. As in Section~\ref{sec:naive-training}, seated rest periods of self-selected duration were permitted after episodes 17 and 34, during which data collection was paused; TF01 completed all episodes continuously, while TF02 and TF03 utilized the rest periods. Per-episode protocol details for both phases are in the supplementary material (Section~\ref{sec:supp-protocol}).

\subsection{Data Analysis}
\label{sec:data-analysis}

Simulator fidelity was quantified by Pearson correlation $r$ between simulation-predicted and hardware-measured returns over the 25 evaluated controllers per participant. Stance-phase biomechanical performance was assessed by root-mean-square error (RMSE) relative to able-bodied references for three metrics per joint: angle tracking, commanded torque, and able-bodied-implied torque. The replay assumption was validated by computing inter-controller standard deviation across the 25 hardware-evaluated controllers, expressed as percentage of joint range of motion (\%ROM) where ROM is the peak-to-peak range of the first controller's mean stance-phase trajectory; the assumption is supported when boundary-condition SD as \%ROM is substantially smaller than controlled-joint SD. Detailed analysis procedures are described in the supplementary material (Section~\ref{sec:supp-protocol}).

\section{Results} \label{Sec:Results}

Simulation-based policy optimization converged for all three participants within 5,000--9,000 episodes, and zero-shot deployment of 25 simulation-selected candidate controllers per participant demonstrated high simulation-to-hardware correlation (Pearson $r > 0.96$, $R^2 > 0.92$; Fig.~\ref{fig:pretrain-correlation}). The hardware-best controller ranked within the top five simulation predictions for all participants. Total-best controllers improved combined stance-phase return by 42\% (TF01), 59\% (TF02), and 55\% (TF03) relative to the unpersonalized HKIC baseline (Table~\ref{tab:hw_total_reward}).

\subsection{Simulation-Based Policy Optimization Convergence}
\label{sec:simulation-training-results}
Policy optimization achieved stable convergence for all participants within approximately 16--29 hours of offline computation (Fig.~\ref{fig:r2-sim-sweep}). Training returns, computed as cumulative episode rewards accumulated during policy updates with exploration noise, converged to stable plateaus by episodes 3,000--5,000 for TF01 and TF02, and by episodes 5,000--9,000 for TF03 (Fig.~\ref{fig:r2-sim-sweep}a). Validation rewards, computed on fixed replay trajectories without exploration noise, exhibited non-monotonic trajectories with two local minima before reaching final convergence (Fig.~\ref{fig:r2-sim-sweep}b). TF01 and TF02 each showed performance troughs near episodes 500--600 and 1,700--2,100 before converging to approximately $-2.9$ and $-2.4$ by episode 3,000, respectively. TF03 exhibited the most complex trajectory with two deep troughs before converging to approximately $-9$ by episodes 5,000--9,000.

\subsection{Zero-Shot Deployment and Simulation-to-Hardware Validation}
\label{sec:pretrain-validation}

\begin{figure*}
  \centering
  \includegraphics[width=\textwidth]{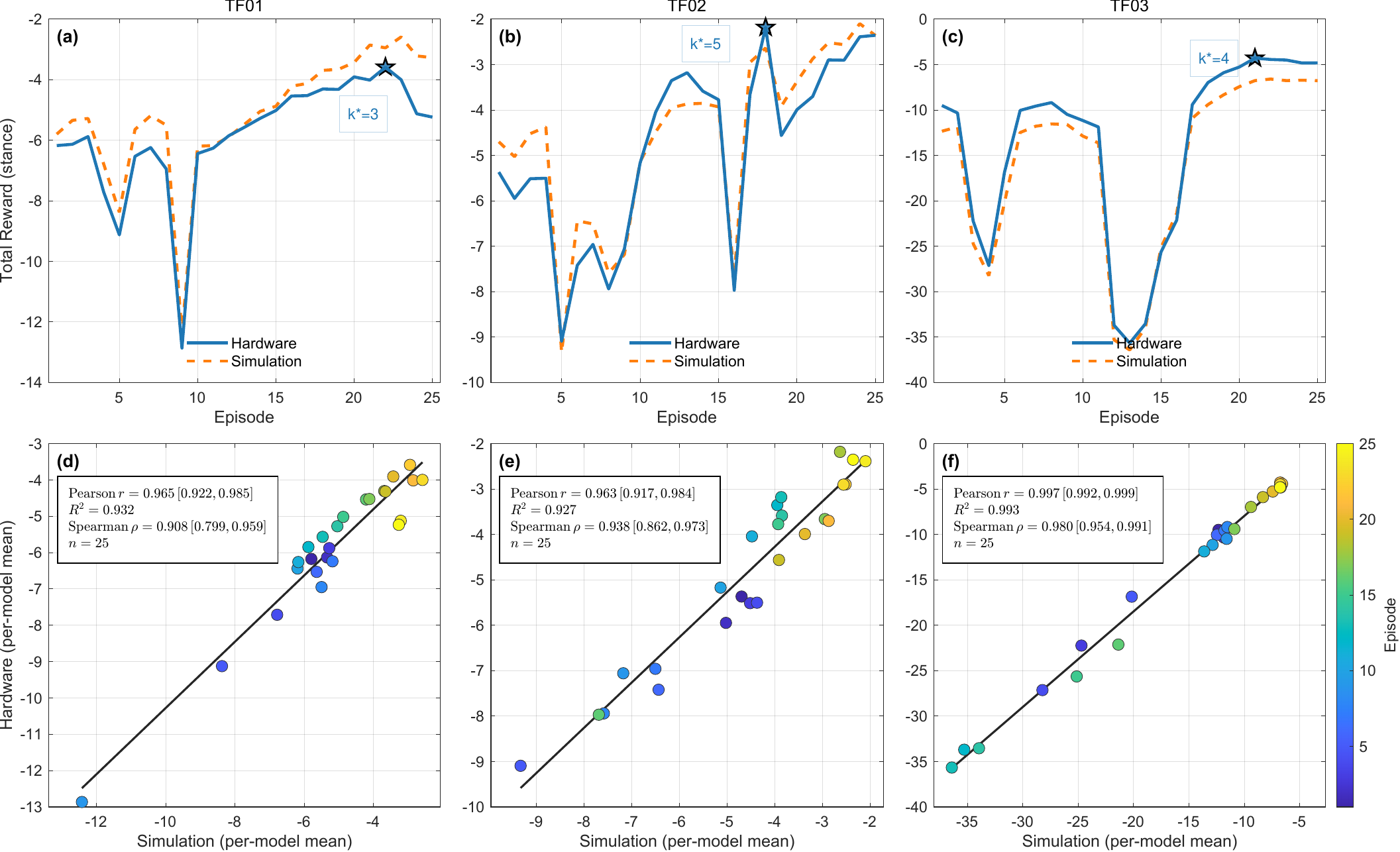}
    \caption{\textbf{Zero-shot deployment of 25 simulation-selected controllers across three participants with transfemoral amputation during Session~2 evaluation.} (\textbf{a--c}) Episode-wise total returns for hardware (blue solid) and simulation (orange dashed) with frozen actor parameters, one controller per episode. Star indicates hardware-best controller; $k^*$ denotes the simulation rank of the hardware-best controller out of over 5,000 or 9,000 simulation candidates. (\textbf{d--f}) Simulation-predicted versus hardware-observed returns, colored by episode number (colorbar, right). Pearson $r$, $R^2$, and Spearman $\rho$ with 95\% confidence intervals reported ($n=25$ per participant). The hardware-best controller appeared within the top five simulation-ranked candidates for all participants ($k^*=3$, $5$, and $4$ for TF01, TF02, and TF03, respectively).}
    \label{fig:pretrain-correlation}
\end{figure*}

Zero-shot deployment of simulation-selected controllers on the powered knee-ankle prosthesis developed in~\cite{elery2020design} demonstrated high correlation between simulation-predicted and hardware-observed returns across all three participants (Fig.~\ref{fig:pretrain-correlation}). TF01 exhibited Pearson $r=0.965$ [0.922, 0.985], Spearman $\rho=0.908$ [0.799, 0.959], and $R^2=0.932$; TF02 exhibited $r=0.963$ [0.917, 0.984], $\rho=0.938$ [0.862, 0.973], and $R^2=0.927$; TF03 exhibited $r=0.997$ [0.992, 0.999], $\rho=0.980$ [0.954, 0.991], and $R^2=0.993$. All correlations were statistically significant ($p<0.001$, $n=25$ controllers per participant).

Simulation rankings reliably identified the hardware-best controller among the 25 candidates evaluated per participant. When controllers were ranked by simulation-predicted performance, the hardware-best controller appeared at rank $k^*=3$ for TF01, $k^*=5$ for TF02, and $k^*=4$ for TF03 (Fig.~\ref{fig:pretrain-correlation}) out of over 5,000 or 9,000 simulation candidates for each participant. Because the 25 candidates were selected using a mixed sampling strategy that spanned both high-validation plateau regions and earlier low-performance training phases, this result demonstrates that simulation correctly filtered out poor-performing controllers drawn from early, unconverged training episodes.

Across the 25 hardware-evaluated controllers, the inter-controller variability of the boundary conditions (thigh angle, foot pitch) remained small in normalized units across the three participants: thigh angle SD was $1.0$--$2.3^\circ$ ($2.7$--$6.5\%$ of joint ROM), and foot pitch SD was $1.0$--$2.7^\circ$ ($3.1$--$5.2\%$ of ROM). The controlled joints (knee and ankle) varied by $2.2$--$5.5^\circ$ ($10.7$--$20.6\%$ of ROM) for knee and $1.1$--$1.8^\circ$ ($10.3$--$11.3\%$ of ROM) for ankle, $3.3\times$ larger on average than the boundary signals in \%ROM units, and consistently larger in every subject and on every joint (per-subject values in Table~\ref{tab:supp-replay-validation}, Fig.~\ref{fig:supp-replay-validation}). Boundary kinematics therefore varied substantially less than the directly controlled joints under impedance perturbation, supporting the replay assumption over the parameter ranges and walking condition explored here.

\begin{table}
\centering
\small
\caption{\textbf{Hardware episode returns (stance phase) for HKIC baseline and representative learned controllers across three participants with transfemoral amputation}. Smaller magnitudes indicate better return; model indices are given in parentheses. Total-best controllers improved combined return by 42\% (TF01), 59\% (TF02), and 55\% (TF03) relative to HKIC baseline. Knee-best and Ankle-best definitions per Table~\ref{tab:r6-complete}.}
\label{tab:hw_total_reward}
\begin{tabular}{lcccc}
\\
\hline
\textbf{Subject} & \textbf{HKIC} & \textbf{Knee-best} & \textbf{Ankle-best} & \textbf{Total-best} \\
\hline
\textbf{TF01} & -6.17 (1) & -12.86 (9) & -5.12 (24) & -3.59 (22) \\
\textbf{TF02} & -5.37 (1) & -2.35 (25) & -4.04 (11) & -2.18 (18) \\
\textbf{TF03} & -9.50 (1) & -33.70 (12) & -10.04 (6) & -4.29 (21) \\
\hline
\end{tabular}
\end{table}

\subsection{Multi-Objective Trade-offs and Learned Impedance Patterns}
\label{sec:tradeoffs-impedance}
Personalized policies improved total stance-phase return by 42--59\% across all participants relative to the unpersonalized HKIC baseline (Table~\ref{tab:hw_total_reward}). No single controller simultaneously minimized knee angle RMSE, ankle angle RMSE, and AB-implied torque RMSE, so Knee-best (lowest knee RMSE), Ankle-best (lowest ankle RMSE), and Total-best (best hardware return) policies corresponded to distinct learned controllers for all three participants (Table~\ref{tab:r6-complete}; phase-specific impedance and joint outcomes at heel strike, mid-stance, and toe-off in Table~\ref{tab:impedance-phase-comparison}).

\begin{figure*}
  \centering
  \includegraphics[width=\textwidth]{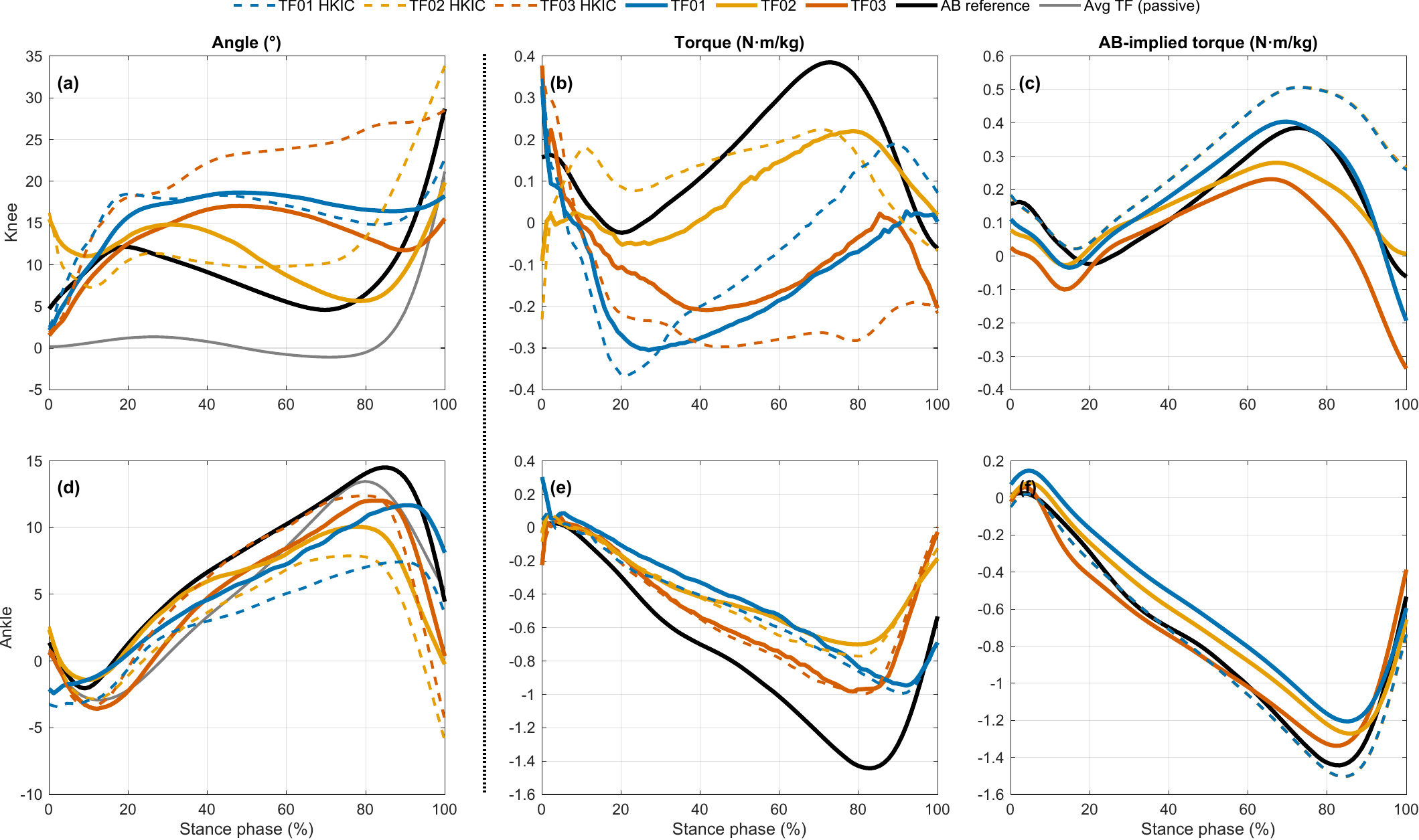}
    \caption{\textbf{Stance-phase joint angle and torque trajectories for Total-best learned controllers versus HKIC baseline across three participants with transfemoral amputation during level-ground walking at 0.8~m/s.} (\textbf{a})~Knee angle. (\textbf{b})~Knee commanded torque. (\textbf{c})~Knee able-bodied-implied torque. (\textbf{d})~Ankle angle. (\textbf{e})~Ankle commanded torque. (\textbf{f})~Ankle able-bodied-implied torque. Light lines indicate HKIC baseline; bold lines indicate Total-best controllers. Line styles: TF01 (blue solid), TF02 (yellow dashed), TF03 (orange dotted). Black solid line: able-bodied reference~\cite{reznick2021lower}. Gray solid line (panels a, d only): average passive prosthesis reference from K3-level prosthesis users~\cite{hood2020kinematic}. Horizontal axis: stance phase (\%). Shaded bands indicate $\pm$1 SD across strides.}
    \label{fig:joint-trajectories-multi-subject}
\end{figure*}

\begin{figure*}
  \centering
  \includegraphics[width=\textwidth]{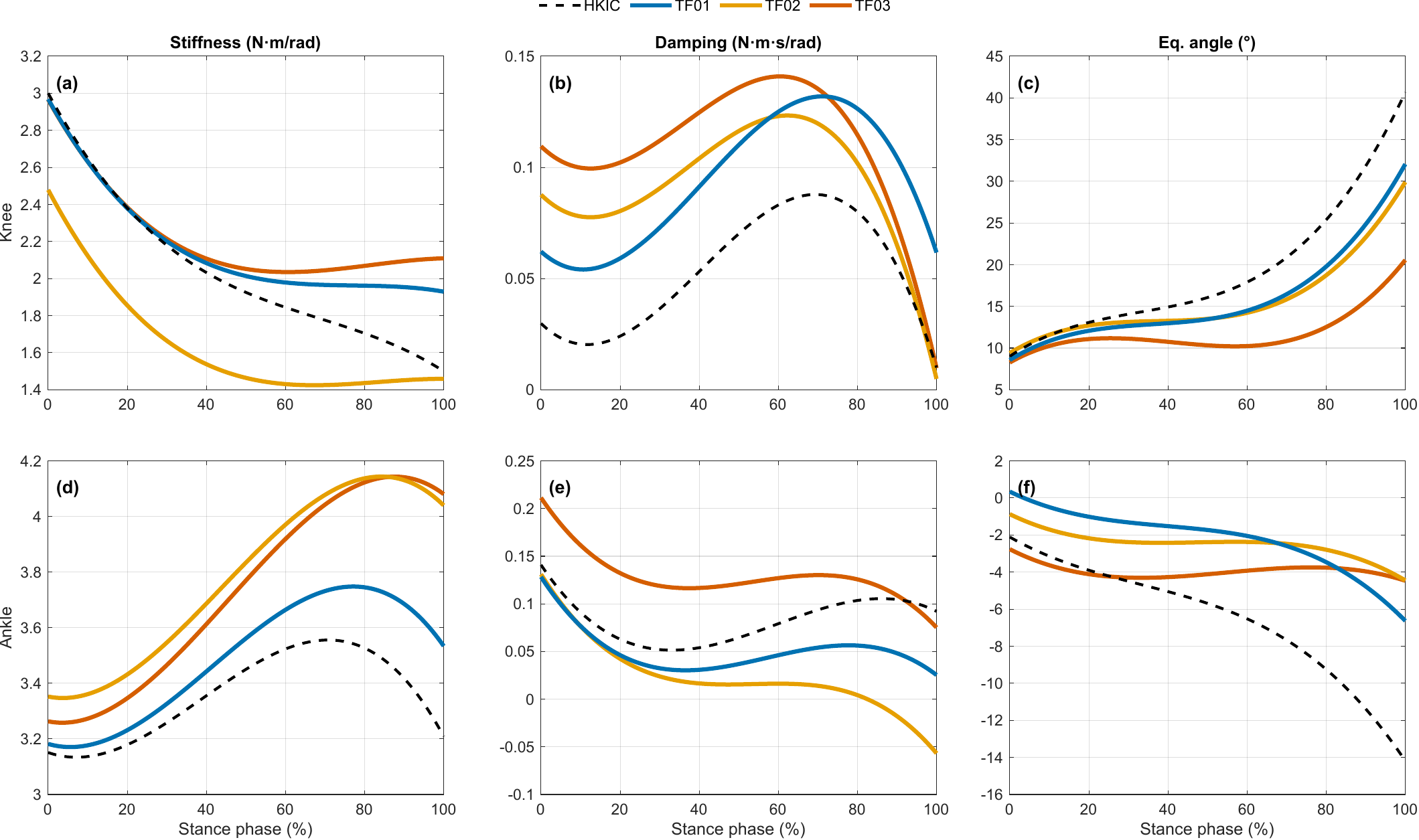}
    \caption{\textbf{Learned phase-dependent impedance parameters for Total-best controllers versus HKIC baseline across three participants with transfemoral amputation.} \textbf{(a)}~Knee stiffness $K$ (N$\cdot$m/rad). \textbf{(b)}~Knee damping $B$ (N$\cdot$m$\cdot$s/rad). \textbf{(c)}~Knee equilibrium angle $\theta^{\mathrm{eq}}$ (deg). \textbf{(d)}~Ankle stiffness $K$ (N$\cdot$m/rad). \textbf{(e)}~Ankle damping $B$ (N$\cdot$m$\cdot$s/rad). \textbf{(f)}~Ankle equilibrium angle $\theta^{\mathrm{eq}}$ (deg). HKIC baseline (black dashed); Total-best controllers: TF01 (blue solid), TF02 (yellow solid), TF03 (orange solid).}
    \label{fig:impedance-multi-subject}
\end{figure*}

Under Total-best controllers, joint angle tracking improved for all participants, with the largest gains at the joint where the HKIC baseline RMSE was largest (per-subject deltas in Table~\ref{tab:r6-complete}; trajectories in Fig.~\ref{fig:joint-trajectories-multi-subject}). The kinematics-versus-torque trade-off was visible in single-objective controllers: Knee-best and Ankle-best policies traded AB-implied torque RMSE (increases of 86--486\%) for improved tracking at their target joint, whereas Total-best controllers achieved more balanced kinetics. Compared to HKIC baseline, personalization changed impedance parameters most at mid-stance and toe-off (knee equilibrium $-$16--49\%, knee damping $+$100--500\%; Fig.~\ref{fig:impedance-multi-subject}; Table~\ref{tab:impedance-phase-comparison}). Full per-subject deltas, Knee-best/Ankle-best detailed RMSE breakdowns, and per-controller impedance curves are provided in the supplementary material (Section~\ref{sec:supp-tradeoffs-detail}, Figs.~\ref{fig:supp-joint-knee-best}--\ref{fig:supp-impedance-ankle-best}).

\subsection{Hardware Fine-Tuning Versus Naïve Training}
\label{sec:finetune-comparison}

Hardware fine-tuning initialized from simulation-optimized controllers maintained performance above HKIC baseline with near-zero Theil--Sen slopes ($m=-0.030$, $-0.030$, $-0.043$ for TF01, TF02, TF03), whereas na\"ive hardware training degraded performance for two of three participants: TF01 returns fell from $-6.2$ to $-10.2$ ($m=-0.103$), TF03 from $-11.1$ to $-14.5$ before discomfort-driven termination at episode~35 ($m=-0.162$), and TF02 from $-4.7$ to $-6.5$ with partial recovery (Figs.~\ref{fig:finetune-comparison},~\ref{fig:supp-naive-training}). Simulation-based optimization was therefore sufficient for identifying high-performing controllers without additional hardware fine-tuning.

\section{Discussion} \label{sec:discussion}

This study establishes that replay-constrained simulators provide sufficient predictive validity for simulation-based controller tuning, enabling full-dimensional personalization of multi-joint prosthesis impedance within clinically practical constraints. Results across three participants with transfemoral amputation demonstrate that encoding human behavior as boundary conditions, rather than explicitly modeling neuromuscular dynamics, offers a viable path to safe, efficient personalization: hardware-best controllers were consistently identified within the top five simulation-ranked candidates, and all participants achieved 42--59\% improvement over the population-derived baseline.

\subsection{Design Choices Underlying Sim-to-Real Transfer}
\label{sec:disc-sim-to-real}

The hardware-best controller appeared within the top five simulation-ranked candidates for all participants and improved stance-phase kinematic tracking relative to the unpersonalized HKIC baseline. This practical success rests on three design choices that constrain the sim-to-real problem rather than attempting to solve it through high-fidelity physiological simulation. First, treating hip kinematics and foot orientation as replay constraints implicitly captures human adaptation without modeling it: boundary-condition variability across 25 hardware-evaluated controllers remained at $2.7$--$6.5\%$ of joint ROM versus $10.3$--$20.6\%$ for the controlled joints, a $3.3\times$ larger normalized excursion (Fig.~\ref{fig:supp-replay-validation}; Table~\ref{tab:supp-replay-validation}), even as learned impedances deviated substantially from HKIC (knee equilibrium $-25$--$53\%$, knee damping up to $+3500\%$). This stability likely reflects feedforward proximal motor planning consistent with the proximal-distal gradient~\cite{daley2007running} for hip kinematics, and the biomechanical requirement of stable ground contact for foot orientation. Second, the physics-consistent policy structure (phase-only actor input, third-order polynomial parameterization) prevents the policy from learning state-dependent reactions that exploit simulation-specific configurations and constrains learned impedance to biologically smooth modulation~\cite{hogan1985impedance}. Third, optimization scope is restricted to stance-phase impedance, where the replay assumption is empirically validated, while preserving the validated HKIC swing controller~\cite{best2023data}.

Two alternative paradigms illustrate the trade-offs the framework navigates. High-fidelity musculoskeletal simulation~\cite{luo2024experiment} optimized a single controller for a hip exoskeleton without individual personalization. Hardware-collected surrogate models~\cite{hong2025offline} enable personalization but require over 30 parameter combinations to train a Gaussian Process Regression surrogate and restrict optimization to a 2-dimensional PCA subspace~\cite{hong2023towards, hong2026efficient}. Against these, the framework concedes that the human side of the human-prosthesis system is difficult to model, and in exchange constrains optimization to the side the simulator captures faithfully. This trade enables full multi-joint continuous impedance personalization from a single three-minute baseline trial, without the predictive neuromuscular modeling, hardware data sweeps, or dimensionality reduction the alternative paradigms require.

\subsection{Multi-Objective Trade-offs and Learned Impedance Adaptations}
\label{sec:tradeoffs-impedance-discussion}

Multi-joint impedance personalization improved overall stance-phase performance but revealed systematic trade-offs among competing objectives rather than universal improvement across all metrics. No single controller simultaneously minimized knee angle RMSE, ankle angle RMSE, and AB-implied torque RMSE, indicating that optimizing for one objective inherently constrains others (Table~\ref{tab:r6-complete}; Fig.~\ref{fig:rmse-controllers}).

\textbf{Sources of multi-objective trade-offs.} The observed trade-offs arise from cross-joint kinematic coupling inherent to serial-chain biomechanics. Controllers achieving minimum knee angle RMSE also reduced ankle angle RMSE but did not achieve the minimum ankle error observed in Ankle-best controllers, and conversely, controllers achieving minimum ankle angle RMSE exhibited variable knee performance (Table~\ref{tab:r6-complete}; Fig.~\ref{fig:rmse-controllers}). This pattern indicates that the knee and ankle are dynamically coupled such that changes in one joint affect the other's performance, supporting the necessity of simultaneous multi-joint personalization. Prior work~\cite{hong2025offline} proposed multi-agent RL for knee-ankle devices, treating each joint as an independent agent; the cross-joint dependencies observed here indicate that unified multi-joint optimization is necessary to capture these interactions. The composite reward function also revealed trade-offs between kinematic tracking and torque agreement: controllers that improved angle tracking adopted impedance parameters that differed substantially from population-derived HKIC values, with equilibrium angle reductions of 25--57\% and damping increases up to 3500\%, increasing AB-implied torque RMSE relative to baseline while keeping commanded torque RMSE largely unchanged.

\textbf{Participant-specific patterns.} HKIC baseline tracking performance varied across participants and joints, and simulation-based personalization provided the greatest benefit at the joint with the largest baseline tracking error (Table~\ref{tab:r6-complete}). TF03 illustrates this pattern most clearly: population-averaged HKIC produced excessive mid-stance knee flexion that the personalized controller corrected toward the able-bodied reference (Fig.~\ref{fig:joint-trajectories-multi-subject}a; deep dive in Supplementary Results, Section~\ref{sec:supp-tradeoffs-detail}). This pattern is consistent with clinical observations that prosthesis users exhibit heterogeneous gait patterns shaped by residual limb anatomy, socket fit, and compensatory strategies~\cite{tucker2015control}, underscoring the need for individualized rather than population-level tuning~\cite{reznick2025clinical}.

\subsection{Clinical Workflow as Demonstration Use-Case}
\label{sec:disc-clinical-workflow}

The framework was designed around three clinical constraints: limited in-clinic time, variable provider infrastructure, and user safety during exploration. Reimbursement compensates device delivery rather than service time, and motion capture~\cite{prasanna2023data}, metabolic measurement~\cite{ingraham2018choosing}, or real-time on-prosthesis inference~\cite{wen2019online, li2021toward} are rarely available outside research settings; existing hardware-based personalization compounds these constraints by exposing users to suboptimal controllers~\cite{wen2019online, alili2023novel}, while existing simulation-based methods achieve safety only through unvalidated dimensionality reduction~\cite{hong2025offline}. The replay-constrained framework addresses each constraint: clinical time reduces to two appointments of approximately three minutes each (baseline collection and top-five candidate evaluation), with the 16--29~hour simulation phase running offline on remote clusters between visits; provider infrastructure simplifies to onboard prosthesis sensors only; and safety is preserved by confining algorithmic exploration to simulation, shielding users from the performance valleys observed near episodes 500--600 and 1{,}700--2{,}100 (Fig.~\ref{fig:r2-sim-sweep}) and avoiding the on-hardware degradation patterns documented in Section~\ref{sec:finetune-comparison}.

\subsection{Limitations and Future Directions}
\label{sec:disc-limitations}

This proof-of-concept validation demonstrates feasibility but not population-level efficacy, and addresses only stance-phase impedance during level-ground walking at 0.8~m/s. HKIC supports variable speeds, inclines, and swing-phase control~\cite{best2023data}; extending the framework requires separate per-condition simulation training and validation that boundary conditions remain stable under expanded control modifications. As only 3 minutes of baseline data is required per condition, the workflow could scale to additional tasks, and multi-task policies could generalize personalized features across conditions~\cite{reznick2025clinical}. The reward function emphasizes biomechanical fidelity rather than clinical outcomes such as metabolic cost~\cite{ingraham2018choosing}, gait symmetry~\cite{wu2021robotic}, or user preference~\cite{clites2021understanding, kim2021influence}; future work could incorporate sound-side IMU measurements or user-reported preferences for individually appropriate rewards.

The replay-constrained simulator required manual tuning of ground reaction force feedback gains (Table~\ref{tab:supp-grf-gains}). TF03 required different values than TF01 or TF02 due to a substantially larger baseline knee-to-ankle RMSE disparity ($13.8^\circ$ versus $3.0^\circ$). Automatic gain adaptation or alternative contact modeling could eliminate this calibration.

\section{Conclusion}
\label{sec:conclusion}

Personalizing stance-phase impedance control for powered knee-ankle prostheses has historically required trade-offs between feasibility and dimensionality. We address this challenge with a \emph{replay-constrained simulation framework}: a MuJoCo-based simulator replays hip kinematics and feedback-augmented ground reaction forces from baseline walking and simulates knee-ankle impedance dynamics in their place, sidestepping the need to predict neuromuscular responses. Training a TD3 policy in this simulator personalizes continuous phase-dependent stiffness, damping, and equilibrium angle at both joints per participant without exposing users to exploration failures. Across three participants with transfemoral amputation, the framework achieved strong simulation-to-hardware predictive validity (hardware-best policies within the top five out of over 5{,}000 or 9{,}000 simulation-ranked candidates) and improved overall rewards by 42--59\% relative to HKIC baseline. The framework establishes feasibility of safe, scalable multi-joint continuous impedance personalization for level-ground walking and is amenable to extension to higher-dimensional controller parameterizations such as neural-network controllers, and to additional locomotion tasks including variable speeds, inclines, and stairs.

\section*{Acknowledgment}
The authors thank Dr.~Curt Laubscher, Andrew Seelhoff, and Albert Lee for their assistance with preliminary system validation, and Prof.~Jennie Si and Hang Liu for valuable discussions on reinforcement learning.

{\footnotesize
\balance
\bibliographystyle{IEEEtran}
\bibliography{bib/references}
}

\newpage

% =========================================================
% SUPPLEMENTARY MATERIAL (one-column)
% =========================================================
\renewcommand{\thefigure}{S\arabic{figure}}
\renewcommand{\thetable}{S\arabic{table}}
\renewcommand{\theequation}{S\arabic{equation}}
\renewcommand{\thepage}{S\arabic{page}}
\renewcommand{\thesection}{S\arabic{section}}
\renewcommand{\thesubsection}{\thesection.\arabic{subsection}}
\setcounter{figure}{0}
\setcounter{table}{0}
\setcounter{equation}{0}
\setcounter{page}{1}
\setcounter{section}{0}
\setcounter{subsection}{0}

\onecolumn

% ---------- Supplement title page ----------
\begin{center}
{\large\bfseries Supplementary Material for}\\[6pt]
{\Large\bfseries A Replay-Constrained Simulation Framework for Personalization of Powered Knee--Ankle Prosthesis Controllers}\\[10pt]
{\normalsize Duong Le, Ryan Posh, Shihao Cheng, Maani Ghaffari, Robert D. Gregg}\\[6pt]
{\small College of Engineering, University of Michigan, Ann Arbor, MI 48109, USA}\\[6pt]
{\small Correspondence: \texttt{duongqle@umich.edu}}
\end{center}

\vspace{12pt}

\begin{figure}[H]
\centering
\begin{minipage}[t]{0.32\linewidth}
\centering
\includegraphics[width=\linewidth]{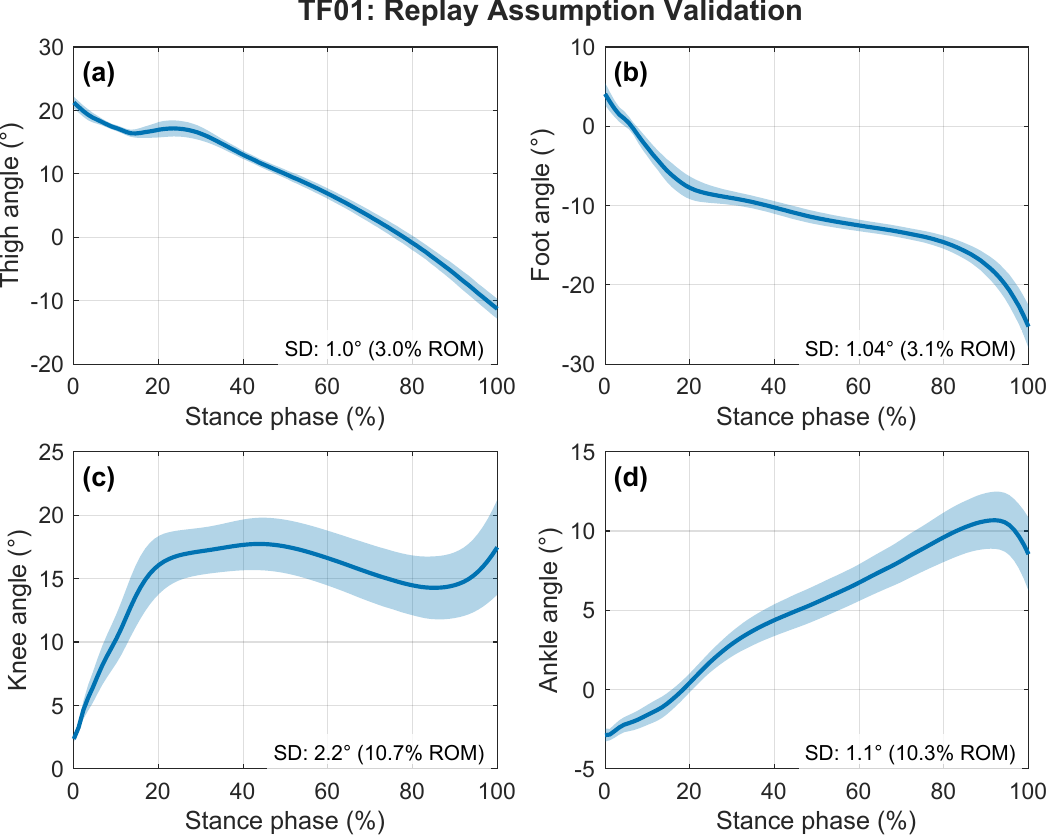}
\textbf{(A)} TF01
\label{fig:supp-replay-TF01}
\end{minipage}
\hfill
\begin{minipage}[t]{0.32\linewidth}
\centering
\includegraphics[width=\linewidth]{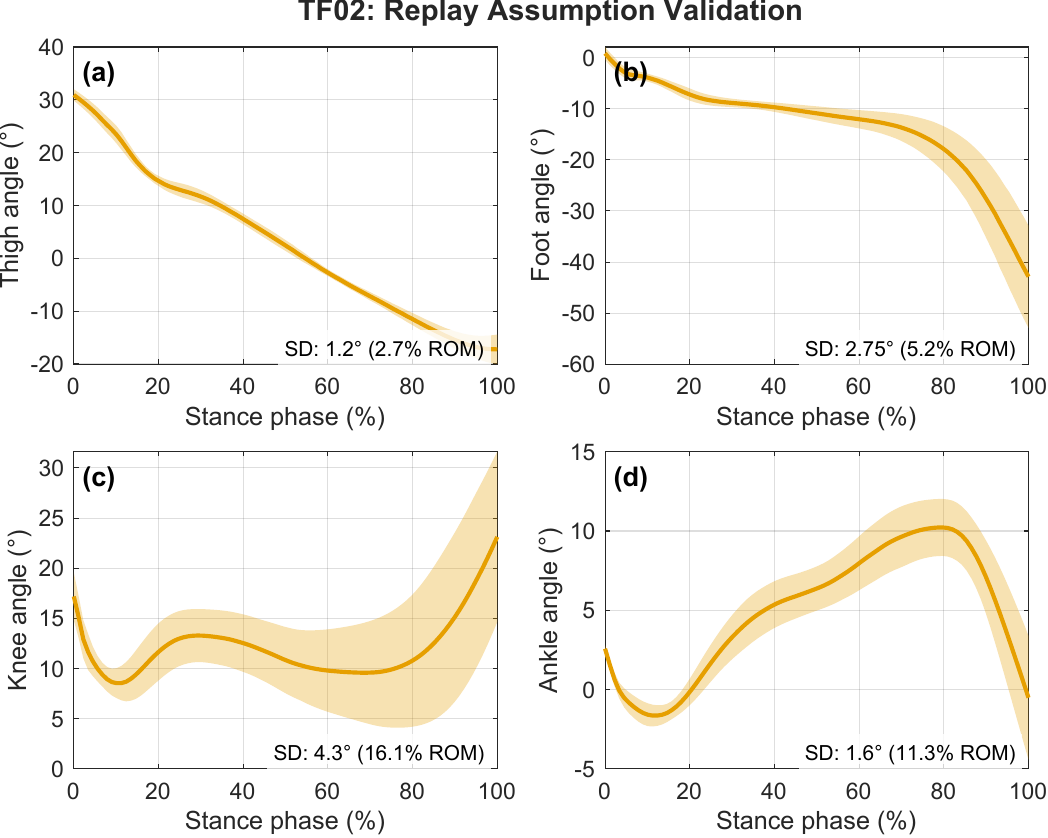}
\textbf{(B)} TF02
\label{fig:supp-replay-TF02}
\end{minipage}
\hfill
\begin{minipage}[t]{0.32\linewidth}
\centering
\includegraphics[width=\linewidth]{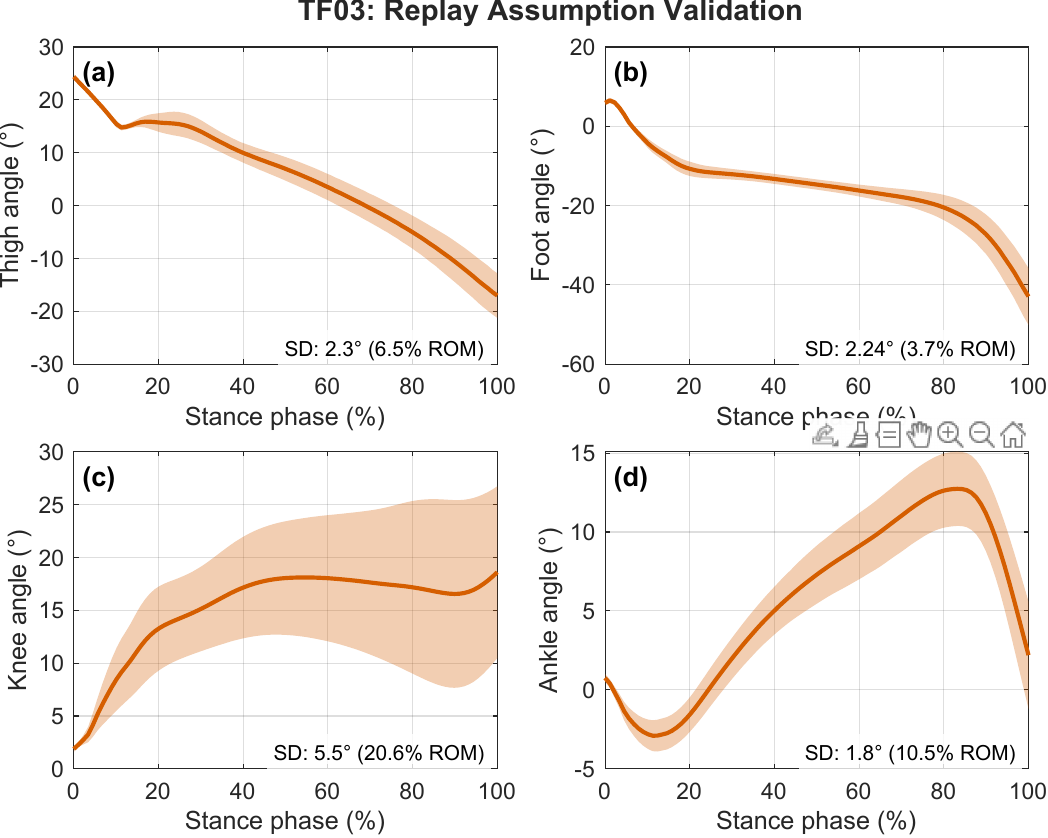}
\textbf{(C)} TF03
\label{fig:supp-replay-TF03}
\end{minipage}
\caption{\textbf{Replay assumption validation: inter-controller variability of boundary conditions versus controlled joints across the 25 hardware-evaluated controllers, for three participants with transfemoral amputation.} (\textbf{A})~TF01. (\textbf{B})~TF02. (\textbf{C})~TF03. Each panel shows \textbf{(a)}~thigh angle, \textbf{(b)}~foot pitch, \textbf{(c)}~knee angle, and \textbf{(d)}~ankle angle. Each controller is first reduced to its mean stance-phase trajectory; the coloured shaded band indicates the inter-controller envelope (across-controller mean $\pm$ SD across the 25 mean curves), and the solid line shows the across-controller mean. Inset numbers report the SD as a percentage of joint range of motion (ROM), where ROM is the peak-to-peak range of the first controller's mean stance trajectory (an in-session reference scale). Boundary conditions (thigh, foot) varied by $2.7$--$6.5\%$ of ROM across all subject $\times$ signal combinations, $3.3\times$ smaller on average than the controlled joints (knee, ankle: $10.3$--$20.6\%$ of ROM) and consistently smaller in every subject, supporting the replay assumption. Per-signal numerical values appear in Table~\ref{tab:supp-replay-validation}.}
\label{fig:supp-replay-validation}
\end{figure}

\begin{figure}[H]
  \centering
  \includegraphics[width=\columnwidth]{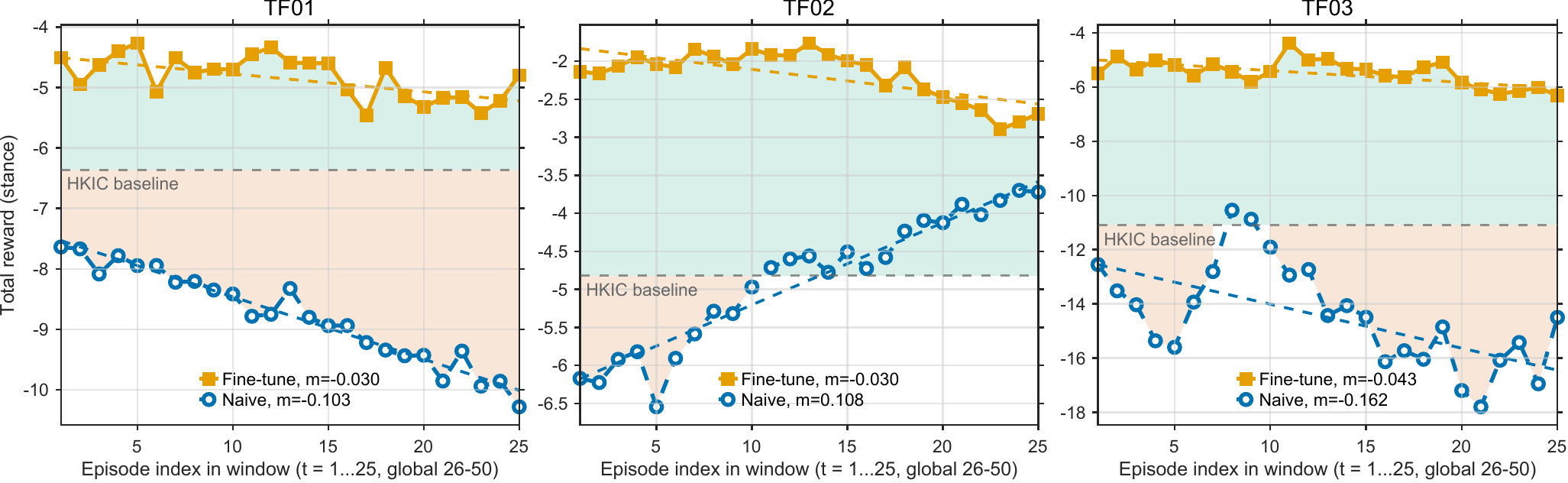}
    \caption{\textbf{Hardware performance during fine-tuning compared to na\"ive training across three participants with transfemoral amputation during level-ground walking at 0.8~m/s.} Orange squares: fine-tuning initialized from hardware-best zero-shot controller. Blue circles: na\"ive training initialized from HKIC baseline. Dashed lines indicate Theil--Sen trends. Fine-tuning exhibited near-zero slopes ($|m| \leq 0.043$); na\"ive training showed continued degradation (TF01, TF03) or recovery (TF02).}
    \label{fig:finetune-comparison}
\end{figure}

\begin{figure}[H]
  \centering
  \includegraphics[width=\linewidth]{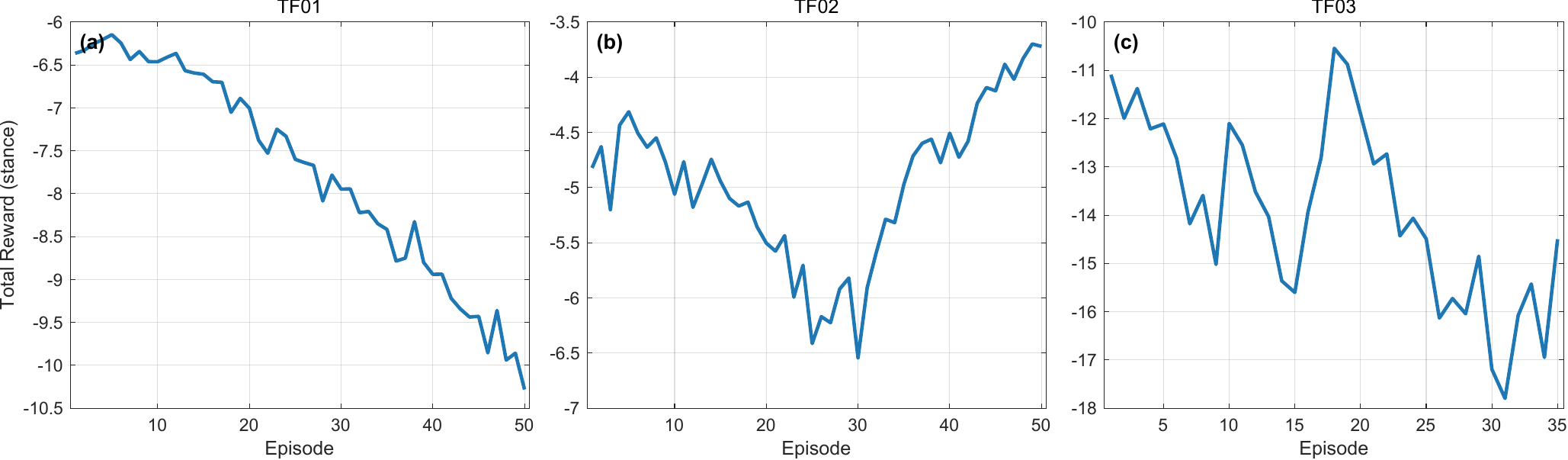}
  \caption{\textbf{Hardware performance during naïve training across three participants with transfemoral amputation during level-ground walking at 0.8~m/s.} Each participant was initialized from HKIC baseline with TD3 policy updates after each seven-stride episode. TF01 and TF03 exhibited monotonic performance degradation over the training session. TF02 demonstrated initial degradation (episodes 1--30) followed by partial recovery (episodes 30--50). Naïve training for TF03 was terminated at episode 35 due to participant-reported discomfort, motivating the simulation-based pre-selection approach to avoid exposing users to exploration-phase instability.}
\label{fig:supp-naive-training}
\end{figure}

\begin{figure}[H]
  \centering
  \includegraphics[width=0.85\linewidth]{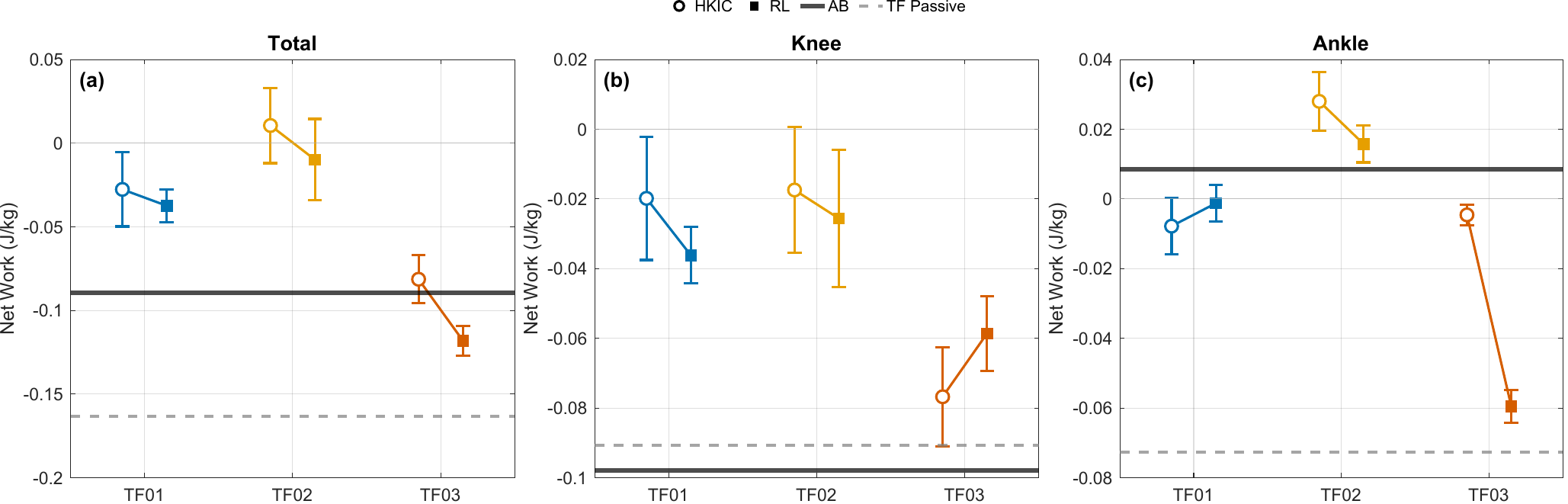}
  \caption{\textbf{Net mechanical work per stride for HKIC baseline and Total-best controllers across three participants with transfemoral amputation during level-ground walking at 0.8~m/s.} \textbf{(a)}~Total (knee + ankle). \textbf{(b)}~Knee. \textbf{(c)}~Ankle. Light bars: HKIC baseline; dark bars: Total-best controllers. Bar colors: TF01 (blue), TF02 (yellow), TF03 (orange). Black solid line: able-bodied reference~\cite{reznick2021lower}. Gray dashed line: passive prosthesis reference~\cite{hood2020kinematic}. Error bars indicate $\pm$1~SD across strides.}
  \label{fig:supp-work-comparison}
\end{figure}

\noindent\textbf{Contents.} This document contains supporting material referenced in the main paper:

\begin{itemize}
\item \textbf{Supplementary Results} (Section~\ref{sec:supp-tradeoffs-detail}): passive-prosthesis reference comparisons, net mechanical work analysis, and energetic interpretation; Fig.~\ref{fig:supp-work-comparison}.
\item \textbf{Supplementary Methods} (Sections~\ref{sec:supp-action-norm}--\ref{sec:supp-protocol}): action-space normalization; full reward formulations (angle tracking, torque agreement, smoothness, damping floor); TD3 implementation and training details; experimental protocol details.
\item \textbf{Supplementary Figures} (Figs.~\ref{fig:supp-replay-validation}--\ref{fig:supp-work-comparison}): replay-assumption validation; simulation-training convergence; per-controller RMSE; per-condition joint trajectories (Knee-best, Ankle-best); per-condition impedance curves; hardware fine-tuning vs.\ na\"ive training; na\"ive hardware training; net mechanical work comparison.
\item \textbf{Supplementary Tables} (Tables~S1--S9): replay-assumption variance summary; GRF feedback gains; participant demographics; impedance parameter bounds; reward weights; observation-space components; TD3 hyperparameters; stance-phase performance metrics; phase-specific impedance and joint outcomes across Knee-best, Ankle-best, and Total-best controllers.
\item \textbf{Supplementary Video} (Video~S1): visualization of the replay-constrained simulator and personalized controllers.
\end{itemize}

\vspace{12pt}
\hrule
\vspace{12pt}

\section{Supplementary Methods}

\subsection{Hip Kinematics Derivation from Contact Constraints}\label{sec:supp-hip-ik}

Expanding on Eq.~\eqref{eq:hip-ik}, $\mathbf{p}_{\mathrm{contact}}^{\mathrm{hip}}$ denotes the contact point position relative to the hip obtained through forward kinematics of the prosthesis kinematic chain at the measured joint configuration. During stance phase, the contact constraint fixes the heel at the world origin ($\mathbf{p}_{\mathrm{heel}} = \mathbf{0}$) for heel contact or the toe at a forward position ($\mathbf{p}_{\mathrm{toe}} = [0, L_{\mathrm{foot}}, 0]^\top$) for toe contact, where $L_{\mathrm{foot}}$ is the foot length and the $y$-axis aligns with the walking direction. Contact type is determined from foot orientation measured by the foot-mounted IMU: heel contact occurs when foot pitch exceeds zero (toe elevated), while toe contact occurs when foot pitch is negative (heel elevated). This binary classification is motivated by normal walking biomechanics: foot orientation maintains non-zero angular velocity throughout stance, passing through flat foot (zero pitch) as a transient zero-crossing rather than a sustained configuration. Consequently, the instantaneous transition between heel and toe contact constraints introduces negligible discontinuity in the reconstructed hip trajectory. Hip velocity is similarly computed by solving the velocity constraint $\dot{\mathbf{p}}_{\mathrm{hip}} = \dot{\mathbf{p}}_{\mathrm{contact}} - \dot{\mathbf{p}}_{\mathrm{contact}}^{\mathrm{hip}}$ with stationary contact constraints ($\dot{\mathbf{p}}_{\mathrm{contact}} = \mathbf{0}$) during stance. This approach reconstructs physically consistent hip trajectories that maintain foot-ground contact throughout stance without requiring hip position sensors.

\subsection{Action Space Normalization}\label{sec:supp-action-norm}

Consistent gradient scaling across impedance parameters with heterogeneous physical units (N$\cdot$m/rad for stiffness, N$\cdot$m$\cdot$s/rad for damping, radians for equilibrium angles) requires normalization. All learning therefore occurs in a normalized action space $[-1,1]^{6}$, with each impedance value normalized via per-parameter affine transformations:
\begin{equation}
a^{\mathrm{norm}}_i = \gamma_i a_i + \delta_i, \quad i \in \{1,\ldots,6\},
\tag{S1}
\label{eq:action-norm}
\end{equation}
where $\gamma_i = 2/(u_i - \ell_i)$ and $\delta_i = -(u_i + \ell_i)/(u_i - \ell_i)$ map physical impedance values $a_i$ to the normalized range $[-1,1]$, with bounds $(\ell_i, u_i)$ derived from hardware actuator limits and physiologically feasible impedance ranges (Table~\ref{tab:supp-limits}). During learning, the actor produces normalized actions with element-wise saturation to $[-1,1]$, and all replay buffer tuples store normalized actions. For torque computation, the inverse transformation maps normalized actions back to physical units: $a^{\mathrm{phys}}_i = \alpha_i a^{\mathrm{norm}}_i + \beta_i$, where $\alpha_i = (u_i - \ell_i)/2$ and $\beta_i = (u_i + \ell_i)/2$.

\subsection{Able-Bodied Gait as Normative Benchmark}

In this work, able-bodied gait serves as a normative benchmark for training and validation rather than a prescriptive target for optimal gait in prosthesis users. This choice aligns with the clinical deployability constraint established in Section~\ref{sec:data-collection}: tracking sound-limb kinematics would require additional sensors on the intact leg~\cite{wu2022reinforcement} or motion capture infrastructure~\cite{prasanna2023data}, whereas able-bodied population averages provide a sensor-independent, reproducible target. Moreover, prior work has demonstrated that restoring normative joint biomechanics in the prosthesis can help restore normative biomechanics to intact joints, reducing compensatory loading patterns and gait asymmetries associated with locomotion in prosthesis users~\cite{zhou2025comparing, best2025clinical}. Although alternative clinical objectives (e.g., metabolic cost~\cite{ingraham2018choosing, kim2021influence}, user preference~\cite{clites2021understanding}) may prove more clinically relevant, rewarding able-bodied kinematic tracking follows established practice in prosthesis control~\cite{gehlhar2023review, liu2025addressing, gao2021reinforcement} and offers a well-defined, reproducible target for validating sim-to-real transfer fidelity before exploring more complex reward formulations.

\subsection{Supplementary Reward Formulations}\label{sec:supp-reward}

\paragraph*{Angle Tracking Term}

The angle tracking term $r^{\theta}_{j,t}$ penalizes normalized kinematic deviations from able-bodied references:
\begin{equation}
r^{\theta}_{j,t} = -w^{\theta}_j \|\tilde{e}^{\theta}_j(t)\|^2_2, \quad \tilde{e}^{\theta}_j(t) = \frac{\theta_j(t) - \theta^{\mathrm{AB}}_j(t)}{\Delta\theta_{j,\mathrm{st}}},
\tag{S1a}
\label{eq:reward_angle}
\end{equation}
where $\theta_j(t)$ is the measured prosthesis joint angle, $\theta^{\mathrm{AB}}_j(t)$ is the able-bodied reference at the current stance phase, and $\Delta\theta_{j,\mathrm{st}}$ is the peak-to-peak range of able-bodied stance-phase data for joint $j$.

\paragraph*{Torque Agreement Term}

Kinematic tracking alone proves insufficient because the policy may converge toward high-stiffness, high-damping solutions with equilibrium angles near able-bodied references, effectively approximating a stiff position controller rather than compliant impedance modulation. The torque agreement term $r^{\tau}_{j,t}$ prevents this by penalizing the difference between able-bodied reference torque and the torque that learned impedance parameters would produce given able-bodied reference kinematics (able-bodied-implied torque):
\begin{equation}
r^{\tau}_{j,t} = -w^{\tau}_j \|\tilde{e}^{\tau}_j(t)\|^2_2, \quad \tilde{e}^{\tau}_j(t) = \frac{\tau^{\mathrm{AB-impl}}_j(t) - \tau^{\mathrm{AB}}_j(t)}{\Delta\tau_{j,\mathrm{st}}},
\tag{S1b}
\label{eq:reward_torque}
\end{equation}
where $\Delta\tau_{j,\mathrm{st}}$ is the peak-to-peak range of able-bodied stance-phase torque for joint $j$. The able-bodied-implied torque is the counterfactual torque that current impedance parameters would generate if driven by able-bodied reference angle $\theta^{\mathrm{AB}}_j(t)$ and velocity $\dot{\theta}^{\mathrm{AB}}_j(t)$:
\begin{equation}
\tau^{\mathrm{AB-impl}}_j(t) = K_j(s_{\mathrm{st}}) \big(\theta^{\mathrm{AB}}_j(t) - \theta^{\mathrm{eq}}_j(s_{\mathrm{st}})\big) + B_j(s_{\mathrm{st}}) \dot{\theta}^{\mathrm{AB}}_j(t).
\tag{S1c}
\label{eq:ab_implied_torque}
\end{equation}
This formulation encourages solutions that preserve HKIC's design philosophy~\cite{best2023data}: producing able-bodied-like torques through moderate impedance gains rather than approximating stiff position control.

\paragraph*{Smoothness Reward Term}

The smoothness term penalizes high-frequency torque variations arising from velocity measurement noise amplified through numerical differentiation:
\begin{equation}
r^{\mathrm{sm}}_{j,t} = -w^{\mathrm{sm}}_j \|\tilde{\mathbf{s}}^{\tau}_j(t)\|^2_2, \quad \tilde{\mathbf{s}}^{\tau}_j(t) = \begin{bmatrix}
\Delta^{(1)}\tau_j(t) / \kappa^{\tau}_j \\
\Delta^{(2)}\tau_j(t) / \kappa^{\tau}_j
\end{bmatrix},
\tag{S2}
\label{eq:reward-smoothness}
\end{equation}
where $\Delta^{(m)}$ denotes the $m$-th finite difference and $\kappa^{\tau}_j = \Delta\tau_{j,\mathrm{st}} / 10$ provides normalization relative to the able-bodied torque range.

\paragraph*{Damping Floor Penalty}

The damping floor penalty discourages negative damping coefficients that would destabilize closed-loop impedance control:
\begin{equation}
r^{\mathrm{damp}}_{j,t} = -w^{\mathrm{damp}}_j H_j(t), \quad H_j(t) = \bigl[\max\{0, -1 - B^{\mathrm{norm}}_{j,\mathrm{pre}}(t)\}\bigr]^2,
\tag{S3}
\label{eq:reward-damping}
\end{equation}
where $B^{\mathrm{norm}}_{j,\mathrm{pre}}(t)$ is the pre-saturation normalized damping action before clipping to the $[-1,1]$ action space bounds. This soft penalty allows the optimizer to explore near-zero damping while discouraging convergence to negative values.

\subsection{TD3 Implementation and Training Details}\label{sec:supp-td3}

\paragraph*{Network Architecture}

TD3 maintains two critic networks $Q_{\theta_1}(o,a)$ and $Q_{\theta_2}(o,a)$ implemented as multilayer perceptrons, each receiving the concatenated observation-action vector $[o_t; a^{\mathrm{norm}}_t] \in \mathbb{R}^{22}$ as input. Each critic consists of three fully connected layers with 256 neurons per layer, using ReLU activation and layer normalization. The actor network performs a linear transformation of the polynomial basis $\boldsymbol{\phi}(s_{\mathrm{st}}) = [1, s_{\mathrm{st}}, s_{\mathrm{st}}^2, s_{\mathrm{st}}^3]^\top$ to output normalized impedance parameters $a^{\mathrm{norm}} \in [-1,1]^6$. Target networks $Q_{\bar{\theta}_1}$, $Q_{\bar{\theta}_2}$, and $\mu_{\bar{\psi}}$ with identical architectures are maintained for stable temporal-difference learning via soft updates.

The actor network receives only $s_{\mathrm{st}} \in [0,1]$ as input, mapping stance phase to continuous impedance trajectories via polynomial evaluation. This architectural choice preserves the HKIC baseline's phase-dependent structure while enabling end-to-end gradient-based optimization. Twin critic networks receive the full observation $o_t \in \mathbb{R}^{16}$ paired with normalized actions $a^{\mathrm{norm}}_t \in [-1,1]^{6}$ for state-action value estimation. To ensure stable learning, the actor is initialized with polynomial coefficients extracted from the HKIC baseline~\cite{best2023data}.

\paragraph*{Training Updates}

During training, critics minimize temporal-difference error using Bellman targets computed from the minimum of two target Q-value estimates with target-policy smoothing, implementing clipped double Q-learning to reduce overestimation bias. The actor updates less frequently than critics (every $d=2$ critic updates) via deterministic policy gradient to maximize $Q_{\theta_1}(o_t, \mu_\psi(s_{\mathrm{st},t}))$. Target networks are soft-updated after each actor update with rate $\tau=5\times10^{-3}$. At each environment step, the agent performs $u$ gradient update steps on both networks, where $u=1$ for simulation training and $u=10$ for hardware sessions. Separate Adam optimizers with learning rate $\eta=1\times10^{-4}$ train actor and critic networks.

\paragraph*{Vectorized Simulation Implementation}

Each training episode employed ten concurrent environment instances: six training environments contributed transitions to the replay buffer with exploration noise ($\sigma_{\mathrm{rollout}}=0.07$), while four validation environments executed deterministic rollouts for continuous performance monitoring without contributing to parameter updates. Each episode consumed ten sequential strides from the shuffled baseline dataset, with training environments accumulating approximately 900 stance-phase transitions before executing a single TD3 parameter update. Convergence was assessed retrospectively by comparing early training returns (episodes 1--500) to the plateau region through linear regression analysis, confirming stable policy performance without overfitting to the limited baseline dataset. All simulations were conducted on a workstation equipped with an NVIDIA GeForce RTX 3090 GPU (24 GB VRAM) using CUDA 13.0.

\subsection{Experimental Protocol Details}\label{sec:supp-protocol}

\paragraph*{Candidate Controller Selection}

We selected 25 candidate controllers per participant from training checkpoints to evaluate simulation-to-hardware prediction fidelity. During training, models achieving new best validation performance were automatically saved whenever validation return improved, creating a sequence of checkpoints spanning the training trajectory. From these automatically-saved checkpoints, we selected controllers using a mixed sampling strategy: approximately 15 controllers were drawn from high-validation plateau regions (episodes where validation performance stabilized near convergence), while the remaining 10 controllers were drawn from earlier training phases, including lower-performance points near validation troughs. This mixed sampling strategy served two purposes: (1) to identify the hardware-best controller among high-performing candidates for clinical deployment, and (2) to probe simulation fidelity across varying policy quality levels, enabling assessment of whether simulation rankings generalized across the full performance spectrum rather than only among near-optimal policies.

The specific checkpoint episodes selected for each participant were:

\begin{itemize}
    \item TF01: $\{0, 18, 34, 300, 600, 900, 1000, 1042, 2100, 2431, 2447, 2466, 2485,$\\
    \hspace{\labelwidth}\phantom{\texttt{type}--} $2514, 2533, 2564, 2581, 2605, 2624, 2652, 2772, 3573, 3605, 4000, 5000\}$
    \item TF02: $\{0, 1, 47, 63, 140, 270, 280, 500, 550, 700, 750, 840, 881,$\\
    \hspace{\labelwidth}\phantom{\texttt{type}--} $935, 974, 1300, 4000, 5000, 2436, 2505, 2595, 2616, 2658, 2769, 2838\}$
    \item TF03: $\{0, 40, 700, 1000, 1300, 1600, 1704, 1793, 1864, 2004, 2060, 2900, 3000,$\\
    \hspace{\labelwidth}\phantom{\texttt{type}--} $3500, 4000, 4300, 5297, 5742, 6365, 7055, 7529, 7589, 7618, 7678, 7707\}$
\end{itemize}

These 25 controllers were drawn from approximately 5,000 training episodes for TF01 and TF02, and 9,000 training episodes for TF03.

% \paragraph*{Hardware-Best Controller Identification ($k^*$)}

% Following hardware evaluation of all 25 candidate controllers, we defined the metric $k^*$ to quantify simulation prediction accuracy. When controllers were ranked by simulation-predicted returns (highest to lowest), $k^*$ denotes the rank position at which the hardware-best controller (the controller achieving the highest return on hardware) appeared in this simulation-based ranking. A value of $k^* = 1$ indicates perfect prediction, meaning the controller that performed best on hardware was also ranked first by simulation out of 5,000 or 9,000 candidates. Lower $k^*$ values indicate more accurate simulation-to-hardware prediction. From a clinical perspective, $k^*$ represents the minimum number of top-ranked simulation candidates that must be tested on hardware to guarantee identification of the hardware-best controller, directly quantifying the hardware evaluation burden required for reliable controller selection.

\paragraph*{Candidate Controller Evaluation with Critic Warm-Up (Episodes 1--25)}
\label{sec:supp-pretrain-eval}

The 25 candidate controllers (Section~\ref{sec:candidate-selection}) were deployed sequentially, with each controller evaluated for one episode. Actor parameters remained frozen at their simulation-optimized values throughout this phase to enable direct assessment of zero-shot policy transfer, while critic networks were updated after each episode using hardware-collected transitions. Critic updates during this phase ensured that when fine-tuning began at episode 26, the critics had accumulated 25 episodes of learning experience, matching the critic training state of the na\"ive baseline (Section~\ref{sec:naive-training}) at the same episode number to enable fair comparison between simulation-guided and na\"ive training trajectories.

\paragraph*{Fine-Tuning with Full Updates (Episodes 26--50)}
\label{sec:supp-finetune-phase}

Following candidate evaluation, HIL fine-tuning proceeded for episodes 26--50. Both actor and critic networks were updated after each episode using full TD3 updates (Section~\ref{sec:rl-framework}), with the participant continuing to walk seven strides per episode at 0.8~m/s. Training hyperparameters were set to $u=10$ gradient steps per episode and policy delay $d=2$, matching the na\"ive baseline to enable direct comparison. The replay buffer contained all stance-phase transitions accumulated from episodes 1--50, enabling the agent to learn from both the initial controller evaluations and ongoing fine-tuning experience.

\paragraph*{Protocol Variation for TF03}

For TF03, technical issues during Session~1 na\"ive training necessitated repeating the na\"ive training protocol during Session~2. Specifically, we conducted na\"ive training at the beginning of Session~2 prior to controller evaluation. This procedural variation is accounted for in the comparative analysis: na\"ive training results for TF03 are compared against fine-tuning performance within the same session rather than across sessions. The simulation-based training and zero-shot controller evaluation proceeded identically to TF01 and TF02.

\section{Supplementary Results}
\label{sec:supp-tradeoffs-detail}

This section provides extended multi-objective trade-off results referenced in Section~\ref{sec:tradeoffs-impedance} of the main text: passive-prosthesis reference comparisons, net mechanical work analysis (Fig.~\ref{fig:supp-work-comparison}), and energetic interpretation.

\subsection{Per-Subject and Per-Controller Trade-off Details}

\begin{figure}[H]
  \centering
  \includegraphics[width=\linewidth]{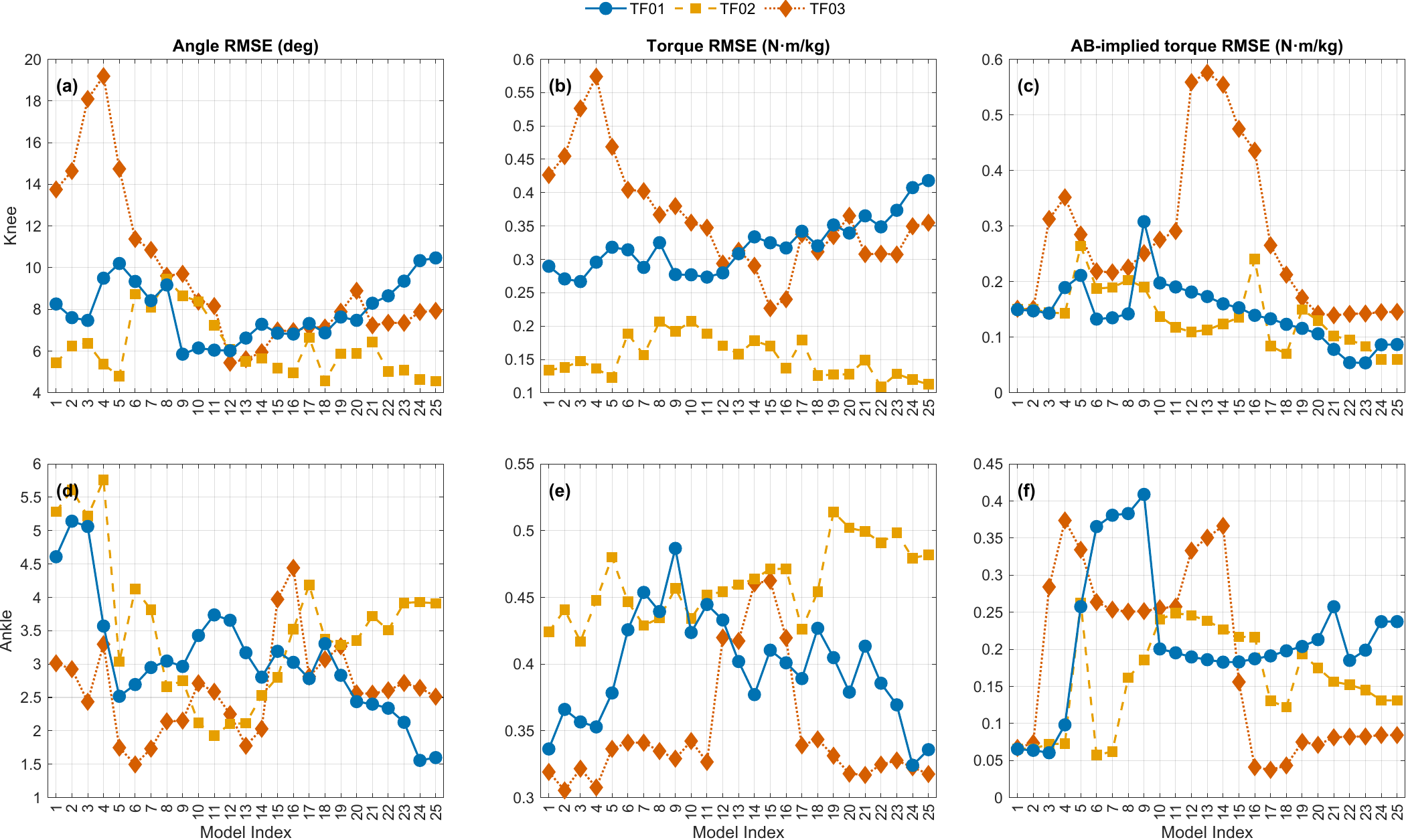}
  \caption{\textbf{Stance-phase RMSE across 25 evaluated controllers for three participants with transfemoral amputation during level-ground walking at 0.8~m/s.} \textbf{(a)} Knee angle RMSE relative to able-bodied reference (deg). \textbf{(b)} Knee commanded torque RMSE relative to able-bodied torque (N$\cdot$m/kg). \textbf{(c)} Knee able-bodied-implied torque RMSE (N$\cdot$m/kg). \textbf{(d)} Ankle angle RMSE relative to able-bodied reference (deg). \textbf{(e)} Ankle commanded torque RMSE relative to able-bodied torque (N$\cdot$m/kg). \textbf{(f)} Ankle able-bodied-implied torque RMSE (N$\cdot$m/kg). Each point represents one controller (mean across strides). Participant markers: TF01 (blue circles), TF02 (yellow squares), TF03 (orange diamonds).}
  \label{fig:rmse-controllers}
\end{figure}

\begin{figure}[H]
  \centering
  \includegraphics[width=\linewidth]{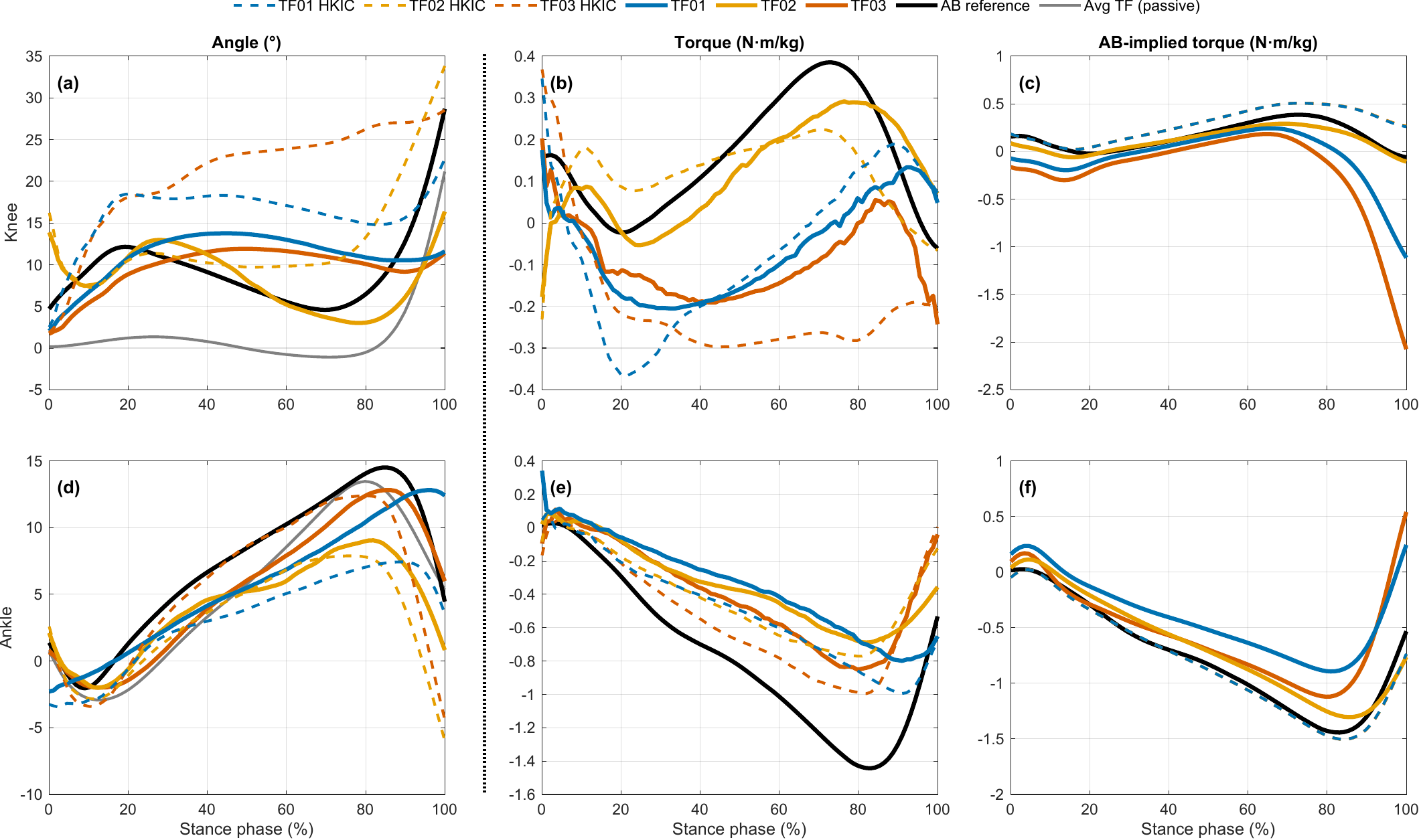}
  \caption{\textbf{Stance-phase joint angle and torque trajectories for Knee-best controllers versus HKIC baseline across three participants with transfemoral amputation during level-ground walking at 0.8~m/s.} Knee-best models: TF01 (model~9), TF02 (model~25), TF03 (model~15). \textbf{(a)}~Knee angle. \textbf{(b)}~Knee commanded torque. \textbf{(c)}~Knee able-bodied-implied torque. \textbf{(d)}~Ankle angle. \textbf{(e)}~Ankle commanded torque. \textbf{(f)}~Ankle able-bodied-implied torque. Light lines indicate HKIC baseline; bold lines indicate Knee-best controllers. Line styles: TF01 (blue solid), TF02 (yellow dashed), TF03 (orange dotted). Black solid line: able-bodied reference. Gray solid line: average passive prosthesis reference from K3-level prosthesis users. Horizontal axis: stance phase (\%).}
  \label{fig:supp-joint-knee-best}
\end{figure}

\begin{figure}[H]
  \centering
  \includegraphics[width=\linewidth]{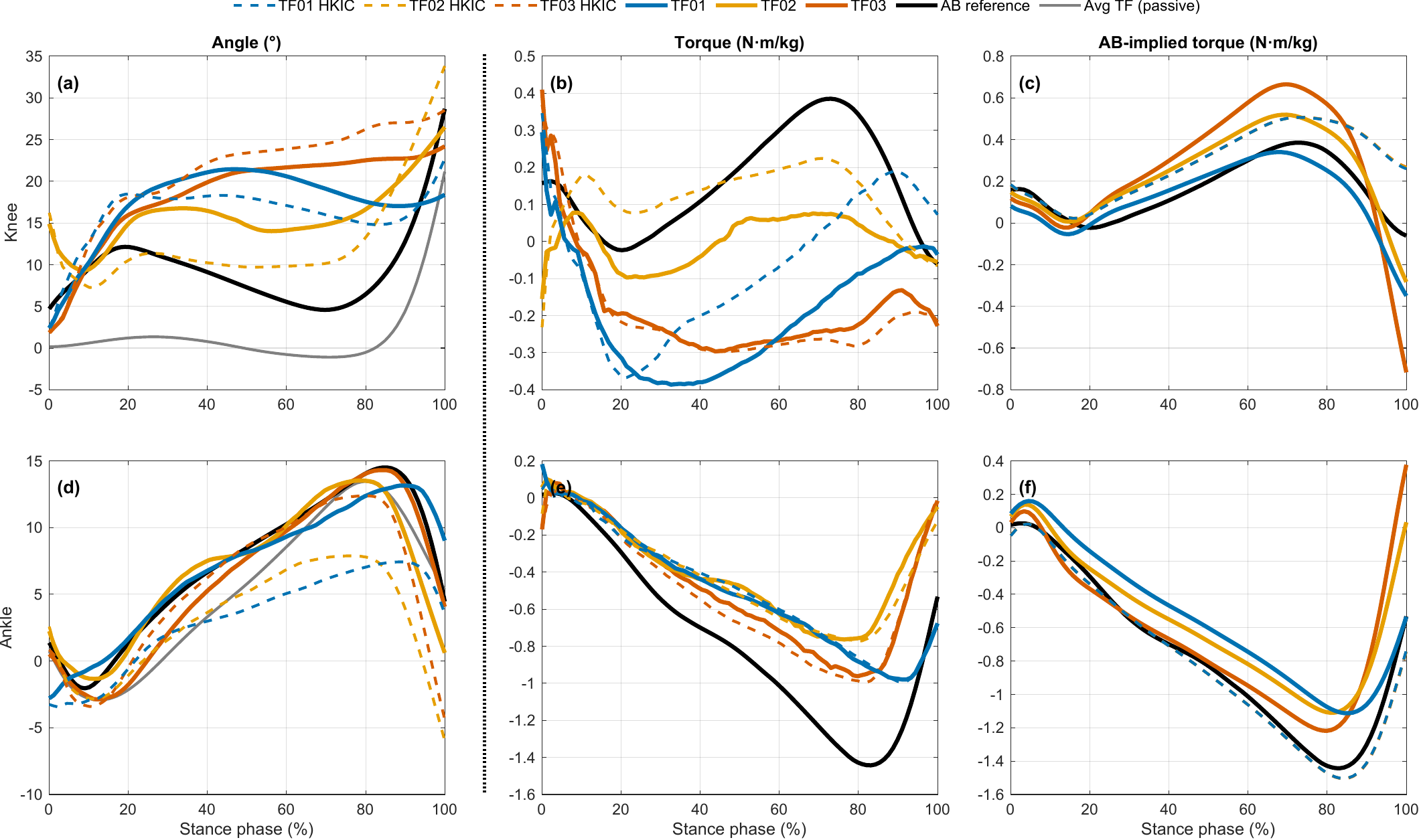}
  \caption{\textbf{Stance-phase joint angle and torque trajectories for Ankle-best controllers versus HKIC baseline across three participants with transfemoral amputation during level-ground walking at 0.8~m/s.} Ankle-best models: TF01 (model~24), TF02 (model~11), TF03 (model~15). \textbf{(a)}~Knee angle. \textbf{(b)}~Knee commanded torque. \textbf{(c)}~Knee able-bodied-implied torque. \textbf{(d)}~Ankle angle. \textbf{(e)}~Ankle commanded torque. \textbf{(f)}~Ankle able-bodied-implied torque. Light lines indicate HKIC baseline; bold lines indicate Ankle-best controllers. Line styles: TF01 (blue solid), TF02 (yellow dashed), TF03 (orange dotted). Black solid line: able-bodied reference. Gray solid line: average passive prosthesis reference from K3-level prosthesis users. Horizontal axis: stance phase (\%).}
  \label{fig:supp-joint-ankle-best}
\end{figure}

Joint angle tracking improvements under Total-best controllers varied across participants: TF01 reduced ankle angle RMSE by 50\% with slight knee degradation (5\% increase); TF02 improved both joints (knee 15\%, ankle 36\%); and TF03 improved primarily at the knee (48\%) with modest ankle improvement (13\%) (Table~\ref{tab:r6-complete}; Fig.~\ref{fig:joint-trajectories-multi-subject}a,d; Fig.~\ref{fig:rmse-controllers}a,d). Knee-best controllers reduced knee angle RMSE by 15--61\% while Ankle-best controllers reduced ankle angle RMSE by 50--65\%, with the other joint always trading off (Table~\ref{tab:r6-complete}; Figs.~\ref{fig:supp-joint-knee-best}--\ref{fig:supp-joint-ankle-best}). HKIC baseline RMSE was a strong predictor of improvement magnitude: TF03's largest knee RMSE ($13.8^\circ$) yielded the largest knee improvement (61\%), and TF02's largest ankle RMSE ($5.3^\circ$) yielded the largest ankle improvement (64\%).

\begin{figure}[H]
  \centering
  \includegraphics[width=\linewidth]{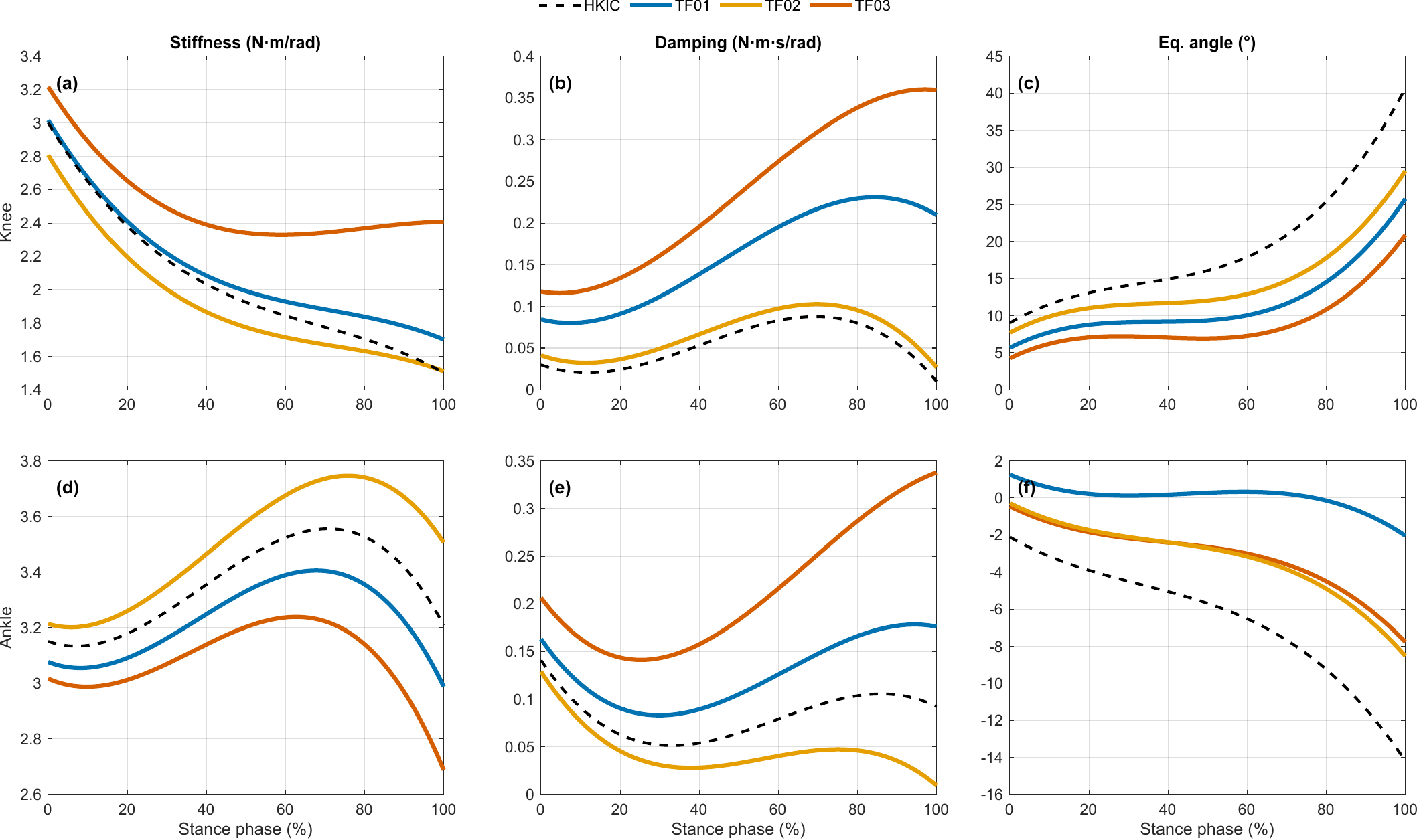}
  \caption{\textbf{Learned phase-dependent impedance parameters for Knee-best controllers versus HKIC baseline across three participants with transfemoral amputation.} Knee-best models: TF01 (model~9), TF02 (model~25), TF03 (model~15). \textbf{(a)}~Knee stiffness $K$ (N$\cdot$m/rad). \textbf{(b)}~Knee damping $B$ (N$\cdot$m$\cdot$s/rad). \textbf{(c)}~Knee equilibrium angle $\theta^{\mathrm{eq}}$ (deg). \textbf{(d)}~Ankle stiffness $K$ (N$\cdot$m/rad). \textbf{(e)}~Ankle damping $B$ (N$\cdot$m$\cdot$s/rad). \textbf{(f)}~Ankle equilibrium angle $\theta^{\mathrm{eq}}$ (deg). HKIC baseline (black dashed), Total-best controllers: TF01 (blue solid), TF02 (yellow solid), TF03 (orange solid).}
  \label{fig:supp-impedance-knee-best}
\end{figure}

\begin{figure}[H]
  \centering
  \includegraphics[width=\linewidth]{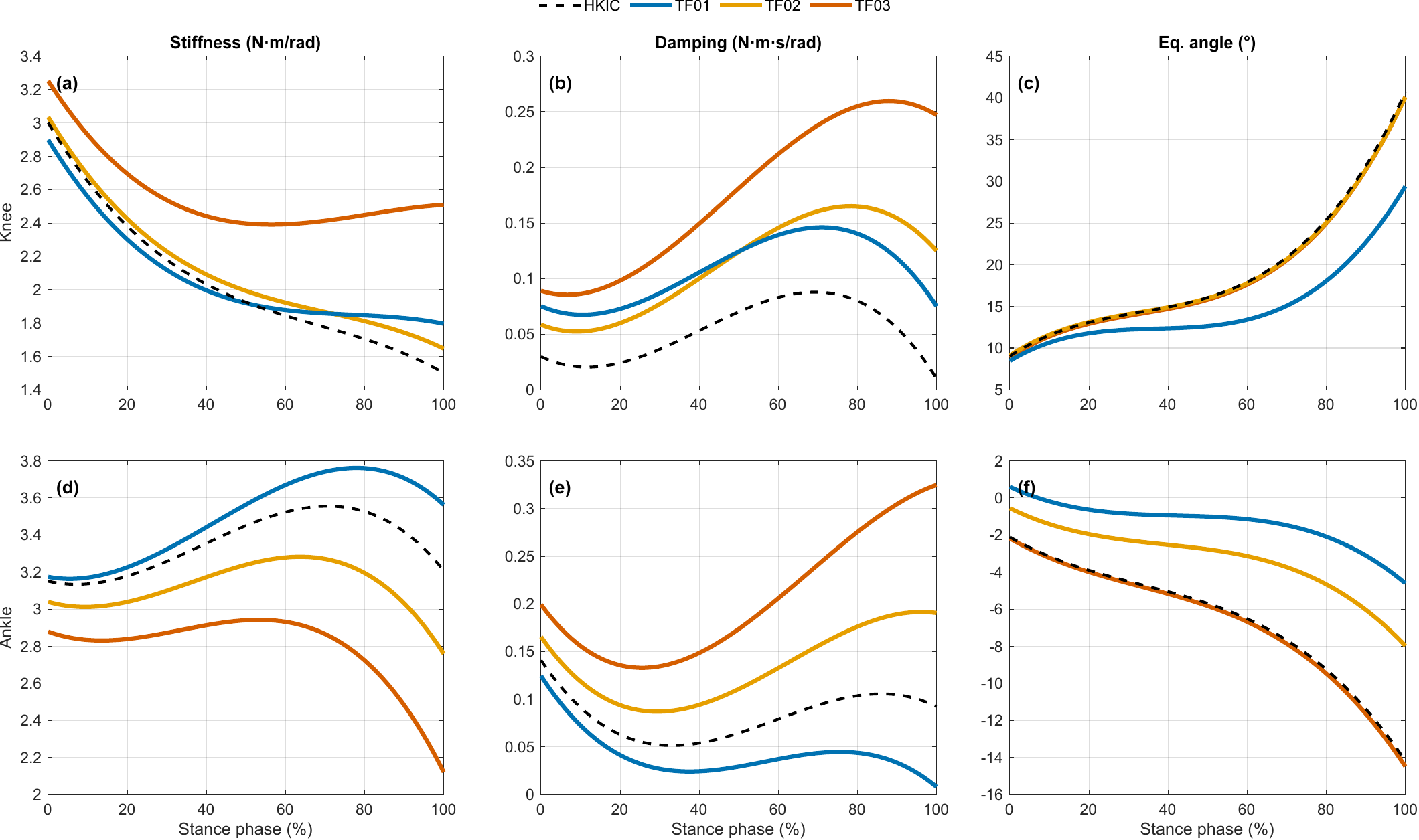}
  \caption{\textbf{Learned phase-dependent impedance parameters for Ankle-best controllers versus HKIC baseline across three participants with transfemoral amputation.} Ankle-best models: TF01 (model~24), TF02 (model~11), TF03 (model~15). \textbf{(a)}~Knee stiffness $K$ (N$\cdot$m/rad). \textbf{(b)}~Knee damping $B$ (N$\cdot$m$\cdot$s/rad). \textbf{(c)}~Knee equilibrium angle $\theta^{\mathrm{eq}}$ (deg). \textbf{(d)}~Ankle stiffness $K$ (N$\cdot$m/rad). \textbf{(e)}~Ankle damping $B$ (N$\cdot$m$\cdot$s/rad). \textbf{(f)}~Ankle equilibrium angle $\theta^{\mathrm{eq}}$ (deg). HKIC baseline (black dashed), Total-best controllers: TF01 (blue solid), TF02 (yellow solid), TF03 (orange solid).}
  \label{fig:supp-impedance-ankle-best}
\end{figure}

Per-subject torque-agreement deltas: Total-best controllers reduced knee AB-implied torque RMSE by 7--67\% across participants while ankle AB-implied torque RMSE increased by 14--157\%. Commanded torque RMSE showed smaller changes than AB-implied torque RMSE across all controllers (Fig.~\ref{fig:rmse-controllers}b,c,e,f; Table~\ref{tab:r6-complete}). Ankle stiffness increased at toe-off by 9--28\% relative to baseline, while heel-strike impedance values remained within 15\% of baseline. Knee-best and Ankle-best controllers exhibited analogous adaptation patterns to Total-best, with full per-controller parameter curves in Figs.~\ref{fig:supp-impedance-knee-best}--\ref{fig:supp-impedance-ankle-best}.

\subsection{Comparison with Passive Prosthesis References}

Both the unpersonalized HKIC baseline and the personalized controllers led to gait more similar to able-bodied references than to references from passive prosthesis users. Average passive prosthesis gait data from K3-level prosthesis users at 0.8~m/s~\cite{hood2020kinematic} shows characteristic flat stance-phase knee angle ($\sim$0$^\circ$; Fig.~\ref{fig:joint-trajectories-multi-subject}a, gray), whereas normative able-bodied gait~\cite{reznick2021lower} exhibits significant early-stance knee flexion to ease weight acceptance. Both baseline and personalized controllers produced dynamic knee flexion throughout stance, including early stance knee flexion, but Total-best personalized controllers further reduced joint angle RMSE relative to able-bodied references (Table~\ref{tab:r6-complete}). Ankle angle trajectories from these powered controllers showed comparable range of motion to both healthy ankle data and passive prosthesis data, as passive ankle-foot prostheses are optimized for level walking kinematics (Fig.~\ref{fig:joint-trajectories-multi-subject}d).

\subsection{Net Mechanical Work}

Net mechanical work per stride varied across participants and control conditions (Fig.~\ref{fig:supp-work-comparison}). Average passive prosthesis knee work ($-0.091$~J/kg) from~\cite{hood2020kinematic} is similar to AB reference ($-0.098$~J/kg) from~\cite{reznick2021lower}, whereas passive ankle work ($-0.073$~J/kg) is substantially more negative than AB ($+0.009$~J/kg), contrasting with the kinematic similarities between these references seen in Fig.~\ref{fig:joint-trajectories-multi-subject}d. On average, HKIC baseline produced less negative total work than AB for all participants: TF01 ($-0.028$~J/kg), TF02 ($+0.011$~J/kg), and TF03 ($-0.081$~J/kg) versus AB ($-0.089$~J/kg). Personalized HKIC (total-best) shifted total work more negative for all participants: TF01 ($-0.037$~J/kg), TF02 ($-0.010$~J/kg), and TF03 ($-0.118$~J/kg), moving TF01 and TF02 closer to the AB reference. Personalization for TF01 and TF02 resulted in knee and ankle work that was consistently more similar to AB references than the HKIC baseline. TF03 exhibited a distinct response: rather than converging toward AB at individual joints, the learned controller redistributed work between knee and ankle. TF03's total work ($-0.118$~J/kg) remained more similar to the AB reference than to the passive reference ($-0.163$~J/kg) after personalization, preserving the powered advantage.

\subsection{Commanded-Torque vs.\ AB-Implied Torque Mechanism}

Despite increased AB-implied torque RMSE, commanded torque RMSE did not increase proportionally and in some cases improved (Table~\ref{tab:r6-complete}). Commanded torque is computed from actual joint angles via the impedance control law (Eq.~\ref{eq:hkic_stance}), so when angle tracking improves such that $\theta$ approaches $\theta^{\mathrm{AB}}$, the resulting commanded torques naturally approach able-bodied references. Improving actual kinematics therefore produces appropriate commanded torques through the impedance law, even when underlying impedance parameters deviate substantially from population-derived values.

\subsection{TF03 Participant Deep Dive}

TF03 illustrates the need for personalization most clearly. Although baseline HKIC deployed on the Össur Power Knee has demonstrated clinical advantages over passive microprocessor knees~\cite{best2025clinical}, population-averaged impedance parameters do not guarantee normative kinematics for every individual. In this work, TF03 exhibited excessive mid-stance knee flexion with baseline HKIC, remaining elevated relative to the able-bodied reference throughout stance rather than extending toward single-limb support (Fig.~\ref{fig:joint-trajectories-multi-subject}a, light orange versus black). While still distinct from passive gait, this deviation from normative kinematics may diminish the intended clinical benefits of a powered prosthesis. The personalized controller reduced this mismatch, decreasing mid-stance knee flexion and bringing the trajectory closer to the able-bodied reference while maintaining clear improvement over passive prosthesis gait (Fig.~\ref{fig:joint-trajectories-multi-subject}a, bold orange versus gray).

\subsection{Energetic Responses (Detailed)}

Net mechanical work analysis reveals differences between passive and powered prostheses not apparent from kinematics alone (Fig.~\ref{fig:supp-work-comparison}). Despite comparable ankle range of motion for level walking (Fig.~\ref{fig:joint-trajectories-multi-subject}d), passive prostheses exhibit a substantial ankle work deficit ($-0.073$ vs $+0.009$~J/kg for AB), reflecting the inability of passive feet to generate active push-off. HKIC baseline addresses this deficit but produces higher total work than AB for all participants, suggesting population-derived impedance parameters may over-power the prosthetic limb. These participant-specific energetic patterns highlight that personalized control may converge to individually appropriate strategies rather than uniform population-level biomimicry.

\section{Supplementary Figures}

\begin{figure}[H]
  \centering
  \includegraphics[width=0.5\linewidth]{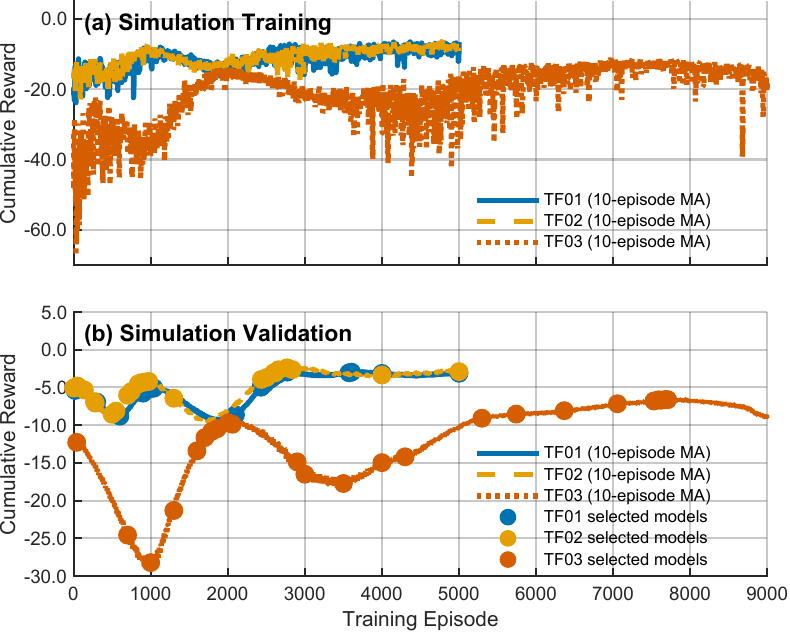}
    \caption{\textbf{Policy optimization over 5,000--9,000 simulation episodes across three transfemoral amputee participants.} \textbf{(a)} Training reward with 10-episode moving averages for TF01 (blue solid), TF02 (yellow dashed), and TF03 (orange dotted), demonstrating convergence by episodes 3,000--5,000 for TF01 and TF02 and by episodes 5,000--9,000 for TF03. \textbf{(b)} Validation reward computed on fixed replay trajectories without exploration noise for each participant. TF01 and TF02 exhibited non-monotonic trajectories with performance troughs near episodes 500--600 and 1,700--2,100 before reaching stable plateaus by episode 3,000. TF03 exhibited a deep performance trough near episode 1,000 (approximately $-28$) and a secondary trough near episode 3,500 (approximately $-18$) before converging by episode 5,000. Circles indicate 25 selected candidate controllers per participant, drawn from both high-validation plateau regions and specific training episodes}
  \label{fig:r2-sim-sweep}
\end{figure}

\clearpage
\section{Supplementary Tables}

\begin{table}
    \centering
    \caption{\textbf{Replay assumption validation: inter-controller SD of boundary conditions and controlled joints across the 25 hardware-evaluated controllers per participant.} For each subject and signal, each controller is first reduced to its mean stance-phase trajectory by averaging across its strides; SD is the per-sample standard deviation across the 25 mean curves, averaged across stance phase. ROM: peak-to-peak range of the first controller's mean stance-phase trajectory (in-session reference scale). Boundary conditions (thigh angle, foot pitch) varied by $2.7$--$6.5\%$ of ROM across all subject $\times$ signal combinations, whereas the controlled joints (knee angle, ankle angle) reached $10.3$--$20.6\%$ of ROM, $3.3\times$ larger on average and consistently larger in every subject, supporting the replay assumption.}
    \label{tab:supp-replay-validation}
    \small
    \begin{tabular}{llccc}
    \hline
    \textbf{Subject} & \textbf{Signal} & SD ($^\circ$) & ROM ($^\circ$) & SD (\% ROM) \\
    \hline
    TF01 & thigh angle & 1.0 & 32 & 3.0 \\
         & foot pitch  & 1.0 & 33 & 3.1 \\
         & knee angle  & 2.2 & 20 & 10.7 \\
         & ankle angle & 1.1 & 11 & 10.3 \\
    \hline
    TF02 & thigh angle & 1.2 & 44 & 2.7 \\
         & foot pitch  & 2.7 & 53 & 5.2 \\
         & knee angle  & 4.3 & 27 & 16.1 \\
         & ankle angle & 1.6 & 14 & 11.3 \\
    \hline
    TF03 & thigh angle & 2.3 & 36 & 6.5 \\
         & foot pitch  & 2.2 & 60 & 3.7 \\
         & knee angle  & 5.5 & 27 & 20.6 \\
         & ankle angle & 1.8 & 17 & 10.5 \\
    \hline
    \end{tabular}
\end{table}

%% GRF Feedback Parameters
\begin{table}
    \centering
    \caption{\textbf{Ground reaction force feedback parameters for level-ground walking. Parameters define the proportional feedback gains ($k_{p,i}$) and mixing coefficients ($\alpha_i$) used in~\eqref{eq:grf-feedback}.}}
    \label{tab:supp-grf-gains}
    \small
    \begin{tabular}{lcccccc}
    \\
    \hline
    \textbf{Subject} & $k_{p,\mathrm{F_x}}$ & $k_{p,\mathrm{F_z}}$ & $k_{p,\mathrm{M}}$ & $\alpha_{\mathrm{F_x}}$ & $\alpha_{\mathrm{F_z}}$ & $\alpha_{\mathrm{M}}$ \\
    \hline
    TF01 & $-5.0$ & $-5.0$ & $-5.0$ & $0.6$ & $0.6$ & $0.8$ \\
    TF02 & $-5.0$ & $-5.0$ & $-5.0$ & $0.6$ & $0.6$ & $0.8$ \\
    TF03 & $-7.0$ & $-5.0$ & $-1.0$ & $1.0$ & $1.0$ & $1.0$ \\
    \hline
    \end{tabular}
\end{table}

%% Participant Characteristics
\begin{table}
\centering
\caption{\textbf{Participant characteristics.} All participants had unilateral transfemoral amputation and were recruited based on Medicare Functional Classification Level K3 or K4. Shank length refers to the total distance from knee joint center to ankle joint center in the prosthesis model.}
\label{tab:supp-participants}
\small
\begin{tabular}{lcccccc}
\hline
\textbf{ID} & \textbf{Sex} & \textbf{Height (cm)} & \textbf{Mass (kg)} & \textbf{Shank (cm)} & \textbf{Side} & \textbf{K-Level} \\
\hline
TF01 & M & 175.0 & 97.0 & 35.6 & R & K4 \\
TF02 & M & 188.9 & 74.7 & 39.4 & L & K4 \\
TF03 & M & 185.4 & 84.0 & 39.1 & R & K4 \\
\hline
\end{tabular}
\end{table}

%% Impedance Parameter Bounds
\begin{table}
    \centering
    \caption{\textbf{Impedance parameter bounds ($\ell_i$, $u_i$) for action space normalization. Units: $K$ in N$\cdot$m/rad, $B$ in N$\cdot$m$\cdot$s/rad, $\theta^{\mathrm{eq}}$ in rad.}}
    \label{tab:supp-limits}
    \small
    \begin{tabular}{lccc}
    \hline
    \textbf{Joint} & \textbf{$K$} [$\ell$, $u$] & \textbf{$B$} [$\ell$, $u$] & \textbf{$\theta^{\mathrm{eq}}$} [$\ell$, $u$] \\
    \hline
    Knee  & [0.5, 5.0] & [0.01, 1.0] & [0, 1.4] \\
    Ankle & [0.5, 5.0] & [0.01, 1.0] & [$-0.7$, 0.7] \\
    \hline
    \end{tabular}
\end{table}

%% Reward Weights
\begin{table}
\centering
\caption{\textbf{Stance-phase reward weights for knee and ankle joints. Global scaling factor $\lambda_{\mathrm{base}} = 20.0$.}}
\label{tab:supp-reward-weights}
\small
\begin{tabular}{lcccc}
\hline
\textbf{Joint} & $w^{\theta}_j$ & $w^{\tau}_j$ & $w^{\mathrm{sm}}_j$ & $w^{\mathrm{damp}}_j$ \\
\hline
Knee  & 1.0 & 1.0 & 0.05 & 2.5 \\
Ankle & 1.0 & 1.0 & 0.4  & 0.1 \\
\hline
\end{tabular}
\end{table}

%% Observation Space Components
\begin{table}
\centering
\caption{\textbf{Observation space components for stance-phase impedance optimization.} The actor network receives only stance-phase progress ($\phi$) as input, while critic networks receive the full 16-dimensional observation.}
\label{tab:supp-observation-space}
\small
\begin{tabular}{llc}
\\
\hline
\textbf{Category} & \textbf{Component} & \textbf{Dim.} \\
\hline
Phase & Stance-phase progress $\phi \in [0,1]$ & 1 \\
\hline
Knee joint & Angle $\theta_{\mathrm{k}}$ & 1 \\
 & Velocity $\dot{\theta}_{\mathrm{k}}$ & 1 \\
 & Able-bodied reference $\theta^{\mathrm{ref}}_{\mathrm{k}}$ & 1 \\
 & Commanded torque $\tau_{\mathrm{k}}$ & 1 \\
\hline
Ankle joint & Angle $\theta_{\mathrm{a}}$ & 1 \\
 & Velocity $\dot{\theta}_{\mathrm{a}}$ & 1 \\
 & Able-bodied reference $\theta^{\mathrm{ref}}_{\mathrm{a}}$ & 1 \\
 & Commanded torque $\tau_{\mathrm{a}}$ & 1 \\
\hline
Thigh IMU & Orientation $\alpha_{\mathrm{th}}$ & 1 \\
 & Angular velocity $\dot{\alpha}_{\mathrm{th}}$ & 1 \\
\hline
Foot IMU & Orientation $\alpha_{\mathrm{ft}}$ & 1 \\
 & Angular velocity $\dot{\alpha}_{\mathrm{ft}}$ & 1 \\
\hline
Load cell & Horizontal force $F_x$ & 1 \\
 & Vertical force $F_z$ & 1 \\
 & Sagittal moment $M_y$ & 1 \\
\hline
\multicolumn{2}{l}{\textbf{Total}} & \textbf{16} \\
\hline
\end{tabular}
\end{table}

%% TD3 Hyperparameters
\begin{table}
\centering
\caption{\textbf{TD3 hyperparameters for stance-phase impedance optimization.}}
\label{tab:supp-hyperparams}
\small
\begin{tabular}{lcc}
\hline
\textbf{Parameter} & \textbf{Simulation} & \textbf{Hardware} \\
\hline
Discount factor ($\gamma$) & 0.99 & 0.99 \\
Target update rate ($\tau$) & $5 \times 10^{-3}$ & $5 \times 10^{-3}$ \\
Target-policy noise std.\ ($\sigma$) & 0.2 & 0.2 \\
Target-policy noise clip ($c$) & 0.5 & 0.5 \\
Policy delay ($d$) & 2 & 2 \\
Gradient steps per episode ($u$) & 1 & 10 \\
Batch size & 512 & 512 \\
Actor learning rate ($\eta_{\pi}$) & $1 \times 10^{-4}$ & $1 \times 10^{-4}$ \\
Critic learning rate ($\eta_{Q}$) & $1 \times 10^{-4}$ & $1 \times 10^{-4}$ \\
Replay buffer size & $5 \times 10^{4}$ & $5 \times 10^{4}$ \\
Learning starts & 5000 steps & 0 steps \\
Rollout noise std.\ ($\sigma_{\mathrm{rollout}}$) & 0.07 & 0 (deterministic) \\
\hline
\end{tabular}
\end{table}

%% Stance-Phase Performance Metrics (RMSE)
\begin{table}
\centering
\caption{\textbf{Stance-phase performance metrics for HKIC baseline and learned controllers across three participants with transfemoral amputation.} Values show RMSE (mean $\pm$ SD across strides) for angle tracking (deg), commanded torque (N$\cdot$m/kg), and able-bodied-implied torque (N$\cdot$m/kg). Four controllers per participant: HKIC baseline (model 1), Knee-best (lowest knee angle RMSE), Ankle-best (lowest ankle angle RMSE), and Total-best (hardware-best from zero-shot evaluation). Numbers in parentheses indicate model index.}
\label{tab:r6-complete}
\footnotesize
\setlength{\tabcolsep}{6pt}
\renewcommand{\arraystretch}{1.15}
\begin{tabular}{@{}llcccccc@{}}
\\
\hline
& &
\multicolumn{2}{c}{\shortstack{\textbf{Angle Tracking RMSE}\\\textbf{vs. Able-Bodied} (deg)}} &
\multicolumn{2}{c}{\shortstack{\textbf{Commanded Torque RMSE}\\\textbf{vs. Able-Bodied} (N$\cdot$m/kg)}} &
\multicolumn{2}{c}{\shortstack{\textbf{AB-implied Torque RMSE}\\\textbf{vs. Able-Bodied} (N$\cdot$m/kg)}} \\
\cline{3-4} \cline{5-6} \cline{7-8}
& & \textbf{Knee} & \textbf{Ankle} & \textbf{Knee} & \textbf{Ankle} & \textbf{Knee} & \textbf{Ankle} \\
\hline

% ================= TF01 ===================
\textbf{TF01} & HKIC (1) 
& $8.3 \pm 1.3$   & $4.6 \pm 0.8$
& $0.29 \pm 0.05$ & $0.34 \pm 0.04$
& $0.15 \pm 0.00$ & $0.07 \pm 0.00$ \\

& Knee-best (9) 
& $5.9 \pm 0.7$   & $3.0 \pm 0.6$
& $0.28 \pm 0.05$ & $0.49 \pm 0.05$
& $0.31 \pm 0.00$ & $0.41 \pm 0.00$ \\

& Ankle-best (24) 
& $10.3 \pm 1.3$   & $1.6 \pm 0.3$
& $0.41 \pm 0.05$ & $0.32 \pm 0.04$
& $0.09 \pm 0.00$ & $0.24 \pm 0.00$ \\

& Total-best (22) 
& $8.7 \pm 0.8$   & $2.3 \pm 0.4$
& $0.35 \pm 0.03$ & $0.39 \pm 0.03$
& $0.05 \pm 0.00$ & $0.18 \pm 0.00$ \\

\hline

% ================= TF02 ===================
\textbf{TF02} & HKIC (1) 
& $5.4 \pm 0.6$   & $5.3 \pm 0.7$
& $0.13 \pm 0.02$ & $0.42 \pm 0.03$
& $0.15 \pm 0.00$ & $0.07 \pm 0.00$ \\

& Knee-best (25) 
& $4.6 \pm 1.4$   & $3.9 \pm 0.6$
& $0.11 \pm 0.01$ & $0.48 \pm 0.02$
& $0.06 \pm 0.00$ & $0.13 \pm 0.00$ \\

& Ankle-best (11) 
& $7.2 \pm 0.4$   & $1.9 \pm 0.3$
& $0.19 \pm 0.01$ & $0.45 \pm 0.02$
& $0.12 \pm 0.00$ & $0.25 \pm 0.00$ \\

& Total-best (18) 
& $4.6 \pm 1.2$   & $3.4 \pm 0.7$
& $0.13 \pm 0.02$ & $0.45 \pm 0.03$
& $0.07 \pm 0.00$ & $0.12 \pm 0.00$ \\

\hline

% ================= TF03 ===================
\textbf{TF03} & HKIC (1) 
& $13.8 \pm 1.6$   & $3.0 \pm 0.3$
& $0.43 \pm 0.05$ & $0.32 \pm 0.02$
& $0.15 \pm 0.00$ & $0.07 \pm 0.00$ \\

& Knee-best (12) 
& $5.4 \pm 0.3$   & $2.2 \pm 0.4$
& $0.29 \pm 0.04$ & $0.42 \pm 0.03$
& $0.56 \pm 0.01$ & $0.33 \pm 0.00$ \\

& Ankle-best (6) 
& $11.4 \pm 0.9$   & $1.5 \pm 0.3$
& $0.40 \pm 0.03$ & $0.34 \pm 0.03$
& $0.22 \pm 0.00$ & $0.26 \pm 0.00$ \\

& Total-best (21) 
& $7.2 \pm 0.5$   & $2.6 \pm 0.2$
& $0.31 \pm 0.03$ & $0.32 \pm 0.02$
& $0.14 \pm 0.00$ & $0.08 \pm 0.00$ \\

\hline

\end{tabular}
\end{table}

%% Phase-Specific Impedance Parameters
\begin{table}
\centering
\caption{\textbf{Impedance parameters and joint outcomes at key stance-phase events across three participants.} Tb: Total-best; Kb: Knee-best; Ab: Ankle-best. $K$: stiffness (N$\cdot$m/rad/kg); $B$: damping (N$\cdot$m$\cdot$s/rad/kg); $\theta^{\mathrm{eq}}$: equilibrium angle (°); $\theta$: prosthesis joint angle (°); $\theta^{\mathrm{AB}}$: able-bodied reference angle (°); $\tau$: joint torque (N$\cdot$m/kg); $\tau^{\mathrm{AB}}$: able-bodied reference torque (N$\cdot$m/kg). Stance phase $s_{\mathrm{st}} \in [0,1]$ corresponds to heel strike (0\%), mid-stance (50\%), and toe-off (100\%).}
\label{tab:impedance-phase-comparison}
\tiny
\setlength{\tabcolsep}{2pt}
\renewcommand{\arraystretch}{1.15}
\begin{tabular}{@{}lllccccccccccccccc@{}}
\\
\hline
& & & \multicolumn{7}{c}{\textbf{Knee}} & \multicolumn{7}{c}{\textbf{Ankle}} \\
\cline{4-10} \cline{11-17}
\textbf{Subj.} & $s_{\mathrm{st}}$ & \textbf{Model} & $K$ & $B$ & $\theta^{\mathrm{eq}}$ & $\theta$ & $\theta^{\mathrm{AB}}$ & $\tau$ & $\tau^{\mathrm{AB}}$ & $K$ & $B$ & $\theta^{\mathrm{eq}}$ & $\theta$ & $\theta^{\mathrm{AB}}$ & $\tau$ & $\tau^{\mathrm{AB}}$ \\
\hline
\textbf{TF01} & 0\% & HKIC & 3.0 & 0.03 & 9.0 & 2.5 & 4.7 & 0.35 & 0.16 & 3.2 & 0.14 & $-$2.1 & $-$3.3 & 1.4 & 0.05 & 0.02 \\
 & & Total Best & 3.0 & 0.06 & 8.6 & 2.1 & 4.7 & 0.33 & 0.16 & 3.2 & 0.13 & 0.3 & $-$2.1 & 1.4 & 0.30 & 0.02 \\
 & & Knee Best & 3.0 & 0.08 & 5.6 & 2.0 & 4.7 & 0.17 & 0.16 & 3.1 & 0.16 & 1.3 & $-$2.3 & 1.4 & 0.34 & 0.02 \\
 & & Ankle Best & 2.9 & 0.08 & 8.5 & 2.4 & 4.7 & 0.29 & 0.16 & 3.2 & 0.12 & 0.6 & $-$2.8 & 1.4 & 0.18 & 0.02 \\
\cline{2-17}
 & 50\% & HKIC & 1.9 & 0.07 & 16.1 & 18.0 & 7.2 & $-$0.14 & 0.21 & 3.5 & 0.06 & $-$5.7 & 4.0 & 8.6 & $-$0.50 & $-$0.84 \\
 & & Total Best & 2.0 & 0.11 & 13.5 & 18.6 & 7.2 & $-$0.23 & 0.21 & 3.6 & 0.04 & $-$1.7 & 6.0 & 8.6 & $-$0.43 & $-$0.84 \\
 & & Knee Best & 2.0 & 0.17 & 9.4 & 13.6 & 7.2 & $-$0.16 & 0.21 & 3.3 & 0.11 & 0.3 & 5.6 & 8.6 & $-$0.33 & $-$0.84 \\
 & & Ankle Best & 1.9 & 0.12 & 12.7 & 21.4 & 7.2 & $-$0.33 & 0.21 & 3.6 & 0.03 & $-$1.0 & 8.2 & 8.6 & $-$0.53 & $-$0.84 \\
\cline{2-17}
 & 100\% & HKIC & 1.5 & 0.01 & 40.7 & 22.7 & 28.7 & 0.07 & $-$0.06 & 3.2 & 0.09 & $-$14.2 & 3.6 & 4.5 & $-$0.68 & $-$0.53 \\
 & & Total Best & 1.9 & 0.06 & 32.1 & 18.2 & 28.7 & 0.00 & $-$0.06 & 3.5 & 0.03 & $-$6.6 & 8.1 & 4.5 & $-$0.69 & $-$0.53 \\
 & & Knee Best & 1.7 & 0.21 & 25.7 & 11.6 & 28.7 & 0.05 & $-$0.06 & 3.0 & 0.18 & $-$2.0 & 12.4 & 4.5 & $-$0.65 & $-$0.53 \\
 & & Ankle Best & 1.8 & 0.08 & 29.4 & 18.4 & 28.7 & $-$0.04 & $-$0.06 & 3.6 & 0.01 & $-$4.6 & 9.0 & 4.5 & $-$0.68 & $-$0.53 \\
\hline
\textbf{TF02} & 0\% & HKIC & 3.0 & 0.03 & 9.0 & 16.2 & 4.7 & $-$0.23 & 0.16 & 3.2 & 0.14 & $-$2.1 & 2.6 & 1.4 & $-$0.09 & 0.02 \\
 & & Total Best & 2.5 & 0.09 & 9.4 & 15.5 & 4.7 & $-$0.09 & 0.16 & 3.4 & 0.13 & $-$0.9 & 2.4 & 1.4 & $-$0.03 & 0.02 \\
 & & Knee Best & 2.8 & 0.04 & 7.7 & 13.9 & 4.7 & $-$0.18 & 0.16 & 3.2 & 0.13 & $-$0.3 & 2.1 & 1.4 & 0.02 & 0.02 \\
 & & Ankle Best & 3.0 & 0.06 & 9.1 & 14.9 & 4.7 & $-$0.16 & 0.16 & 3.0 & 0.17 & $-$0.6 & 2.2 & 1.4 & 0.06 & 0.02 \\
\cline{2-17}
 & 50\% & HKIC & 1.9 & 0.07 & 16.1 & 9.7 & 7.2 & 0.17 & 0.21 & 3.5 & 0.06 & $-$5.7 & 5.3 & 8.6 & $-$0.53 & $-$0.84 \\
 & & Total Best & 1.5 & 0.12 & 13.5 & 11.5 & 7.2 & 0.09 & 0.21 & 3.8 & 0.02 & $-$2.4 & 7.0 & 8.6 & $-$0.48 & $-$0.84 \\
 & & Knee Best & 1.8 & 0.08 & 12.1 & 7.9 & 7.2 & 0.14 & 0.21 & 3.6 & 0.03 & $-$2.7 & 5.3 & 8.6 & $-$0.38 & $-$0.84 \\
 & & Ankle Best & 2.0 & 0.12 & 16.0 & 14.6 & 7.2 & 0.05 & 0.21 & 3.2 & 0.11 & $-$2.8 & 8.2 & 8.6 & $-$0.48 & $-$0.84 \\
\cline{2-17}
 & 100\% & HKIC & 1.5 & 0.01 & 40.7 & 33.8 & 28.7 & $-$0.07 & $-$0.06 & 3.2 & 0.09 & $-$14.2 & $-$6.0 & 4.5 & $-$0.13 & $-$0.53 \\
 & & Total Best & 1.5 & 0.00 & 29.9 & 19.9 & 28.7 & 0.02 & $-$0.06 & 4.0 & -0.06 & $-$4.4 & $-$0.2 & 4.5 & $-$0.18 & $-$0.53 \\
 & & Knee Best & 1.5 & 0.03 & 29.5 & 16.4 & 28.7 & 0.07 & $-$0.06 & 3.5 & 0.01 & $-$8.5 & 0.8 & 4.5 & $-$0.36 & $-$0.53 \\
 & & Ankle Best & 1.6 & 0.12 & 40.1 & 26.5 & 28.7 & $-$0.06 & $-$0.06 & 2.8 & 0.19 & $-$8.0 & 0.6 & 4.5 & $-$0.05 & $-$0.53 \\
\hline
\textbf{TF03} & 0\% & HKIC & 3.0 & 0.03 & 9.0 & 1.9 & 4.7 & 0.37 & 0.16 & 3.2 & 0.14 & $-$2.1 & 0.8 & 1.4 & $-$0.17 & 0.02 \\
 & & Total Best & 3.0 & 0.11 & 8.3 & 1.5 & 4.7 & 0.38 & 0.16 & 3.3 & 0.21 & $-$2.8 & 0.6 & 1.4 & $-$0.23 & 0.02 \\
 & & Knee Best & 3.2 & 0.12 & 4.2 & 1.7 & 4.7 & 0.20 & 0.16 & 3.0 & 0.21 & $-$0.5 & 0.7 & 1.4 & $-$0.10 & 0.02 \\
 & & Ankle Best & 3.3 & 0.09 & 8.9 & 1.8 & 4.7 & 0.41 & 0.16 & 2.9 & 0.20 & $-$2.2 & 0.5 & 1.4 & $-$0.17 & 0.02 \\
\cline{2-17}
 & 50\% & HKIC & 1.9 & 0.07 & 16.1 & 23.4 & 7.2 & $-$0.29 & 0.21 & 3.5 & 0.06 & $-$5.7 & 8.7 & 8.6 & $-$0.68 & $-$0.84 \\
 & & Total Best & 2.1 & 0.14 & 10.3 & 17.0 & 7.2 & $-$0.20 & 0.21 & 3.8 & 0.12 & $-$4.1 & 6.9 & 8.6 & $-$0.65 & $-$0.84 \\
 & & Knee Best & 2.3 & 0.24 & 6.9 & 11.9 & 7.2 & $-$0.17 & 0.21 & 3.2 & 0.18 & $-$2.7 & 6.2 & 8.6 & $-$0.47 & $-$0.84 \\
 & & Ankle Best & 2.4 & 0.18 & 15.8 & 21.3 & 7.2 & $-$0.28 & 0.21 & 2.9 & 0.17 & $-$5.9 & 7.9 & 8.6 & $-$0.61 & $-$0.84 \\
\cline{2-17}
 & 100\% & HKIC & 1.5 & 0.01 & 40.7 & 28.5 & 28.7 & $-$0.22 & $-$0.06 & 3.2 & 0.09 & $-$14.2 & $-$4.4 & 4.5 & 0.00 & $-$0.53 \\
 & & Total Best & 2.1 & 0.01 & 20.6 & 15.5 & 28.7 & $-$0.20 & $-$0.06 & 4.1 & 0.08 & $-$4.5 & 0.4 & 4.5 & $-$0.03 & $-$0.53 \\
 & & Knee Best & 2.4 & 0.36 & 20.9 & 11.4 & 28.7 & $-$0.24 & $-$0.06 & 2.7 & 0.34 & $-$7.8 & 6.0 & 4.5 & $-$0.04 & $-$0.53 \\
 & & Ankle Best & 2.5 & 0.25 & 40.1 & 24.2 & 28.7 & $-$0.23 & $-$0.06 & 2.1 & 0.32 & $-$14.5 & 4.1 & 4.5 & $-$0.02 & $-$0.53 \\
\hline
\end{tabular}
\end{table}

\clearpage
\section{Supplementary Video}

\textbf{Video S1. Demonstration of simulation-based impedance personalization for powered knee-ankle prostheses.} The video presents the replay-constrained simulation framework, walking comparisons between HKIC baseline and learned controllers, and simulation-to-hardware validation results across three participants with transfemoral amputation. The video is available at \url{https://youtu.be/qjFDCWu8QsQ}.

\end{document}